\documentclass[11pt]{article}

\usepackage{acl}
\usepackage[bottom]{footmisc} 

\usepackage{longtable}
\usepackage{array}
\usepackage{booktabs}
\usepackage{ragged2e}
\usepackage{caption}
\usepackage{subcaption}
\usepackage{capt-of}
\usepackage{times}
\usepackage{latexsym}
\usepackage[T1]{fontenc}
\usepackage[utf8]{inputenc}

\usepackage{microtype}
\usepackage{inconsolata}
\usepackage{graphicx}
\usepackage{xcolor}
\usepackage{adjustbox}
\usepackage{float}
\usepackage{makecell}
\usepackage{multirow}
\usepackage{amssymb}
\usepackage{amsmath}
\usepackage{listings}
\usepackage{fvextra}

\DefineVerbatimEnvironment{promptblock}{Verbatim}{
  breaklines=true,
  breakanywhere=true,
  fontsize=\footnotesize,
}

\usepackage{tabularx}
\newcolumntype{Y}{>{\raggedright\arraybackslash}X}

\title{Beyond Surface Forms: A Comprehensive, Mechanism-Oriented Taxonomy
of Indirect Linguistic Encoding for LLM-Based Coded Language Detection%
\rlap{\thanks{Published in the Proceedings of EMNLP 2026
(Main Conference). The official version is available at
\url{https://aclanthology.org/2026.emnlp-main.2444/}.}}}

\author{
Hamid Reza Firoozfar$^{1}$,
Mohammadsadegh Abolhasani$^{1}$, \\
\bfseries Reza Mousavi$^{2}$,
Paul Jen-Hwa Hu$^{1}$ \\
$^{1}$University of Utah, $^{2}$University of Virginia \\
\small Correspondence: \href{mailto:hamid.firoozfar@eccles.utah.edu}{hamid.firoozfar@eccles.utah.edu}
}

\begin{document}
\maketitle
\raggedbottom

\begin{abstract}
To avoid moderation and surveillance on social media, some users routinely invent indirect linguistic expressions (ILE) that camouflage sensitive meanings. Such disguised expressions surface as algospeak, euphemisms, and adversarial obfuscation, depending on intent and context, and often involve recurring encoding mechanisms. We propose a comprehensive, mechanism-oriented taxonomy of ILE that abstracts away from communicative goals and instead categorizes the underlying operations through which meaning is encoded and recovered. We evaluate the taxonomy by incorporating it into large language model (LLM) prompts and comparing it with four existing taxonomies and a no-taxonomy baseline, using 2{,}000 manually annotated TikTok and Bluesky posts. The proposed taxonomy attains the strongest document- and span-level performance across three LLMs. Compared with the best-performing benchmark, its document-level gains are +3.8 accuracy/+4.3 macro-F1 percentage points for GPT-5.4, +0.9/+1.3 for Sonnet 4.6, and +2.1/+2.2 for DeepSeek V4. While the direction of improvement is consistent, the magnitude is model-dependent. The span-level gains are mainly driven by higher recall, sometimes at the cost of precision. In an additional cross-lingual evaluation of 800 Chinese coded-language reviews, the unadapted taxonomy also achieves the highest document- and span-level F1 across the three LLMs. The empirical results reveal the importance of a comprehensive, mechanism-oriented taxonomy as a stable scaffold for detecting emerging coded language and a useful input to content moderation.

\noindent\textit{\textcolor{red}{\textbf{Disclaimer:} This paper contains content that may be profane, vulgar, or offensive.}}
\end{abstract}

\section{Introduction}
\label{sec:intro}

Social media users routinely camouflage sensitive meanings to evade or contest algorithmic moderation and visibility restrictions \citep{klug2023algorithm, steen2023you, calhoun2023they}. Existing forms span a wide spectrum of strategies that we collectively refer to as \textit{indirect linguistic encoding} (ILE). ILE is a moving target for content moderation because users can replace euphemisms and codewords once they become recognizable \citep{zhu2021self, takuro2020codewords}, and detectors built around fixed inventories often struggle with emergent forms \citep{li2025impromptu, sasse2025making}.

\begin{figure}[t]
\centering
\includegraphics[width=0.85\columnwidth]{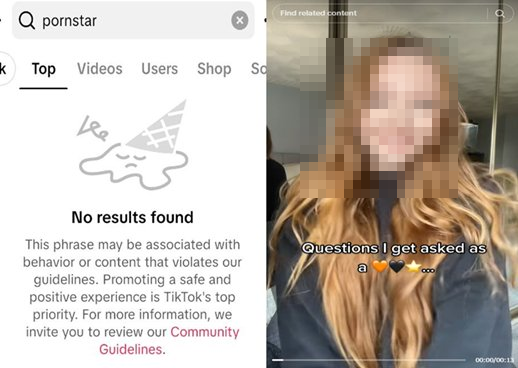}
\caption{An ILE example from TikTok. Left: the flagged term ``pornstar'' returns no results. Right: a creator encodes \textit{porn star} with three emojis, composing three mechanisms (referential iconicity, attribute-based alias, and name-based pictographic mapping).}
\label{fig:example}
\end{figure}

Large language models (LLMs) offer a promising abstraction for ILE detection because they can model context and indirect meaning without explicit lexicons \citep{fillies2024simple}. However, their performance is highly prompt-sensitive \citep{zhuo-etal-2024-prosa}. Recent zero-shot classification studies show that label descriptions and hierarchy-aware guidance can improve classification when labeled examples are unavailable \citep{xia2025ensembling, zhang2025hierprompt, gao2023benefits, kumar2023gen}. These findings motivate testing whether mechanism descriptions can similarly aid ILE detection.

Previous research has established useful taxonomies and inventories for ILE, but each was developed for a particular language, platform, or downstream task and uses a different level of abstraction \citep{calhoun2023they, fillies2024simple, leal2024assessing, zhang2014appropriate}. In practice, users draw on a wide range of strategies and sometimes combine several within a single expression, as illustrated in Figure~\ref{fig:example}. Approaches centered on surface-form perturbation alone (\textit{coc@!ne} for \textit{cocaine}) capture only part of the phenomenon, motivating a taxonomy with broader mechanism coverage for LLM-based detection.

This study investigates how to build a taxonomy of ILE with broader strategy coverage and whether its use improves LLM-based detection. Our contributions are threefold. First, we develop a \textit{mechanism-oriented taxonomy} that organizes encodings by how meaning is concealed and recovered, introducing top-level classes and sub-classes absent from existing schemes. Second, we construct a manually annotated dataset of 2{,}000 recent social media posts (600 Bluesky, 1{,}400 TikTok), labeled by two trained annotators for (i) document-level ILE presence, (ii) all ILE evidence spans, and (iii) each span's mechanism class and sub-mechanism.\footnote{Codes and the publicly releasable portion of the data are available at: \url{https://github.com/hmdfiroozfar/mechanism-oriented-ile-taxonomy}. TikTok-derived text is not redistributed because of data-sharing restrictions.} Third, we compare the proposed taxonomy with four existing taxonomies and a no-taxonomy baseline across three LLMs at both document and span levels, then evaluate all variants on independently annotated Chinese coded-language data, thus providing empirical evidence that the effectiveness of taxonomy-guided prompting depends on the breadth of mechanism coverage rather than the mere use of a taxonomy. Taxonomies with partial coverage can underperform a no-taxonomy prompt, whereas the proposed taxonomy achieves the strongest overall performance, with gains varying across models and settings.

\section{Related Work}
\label{sec:related}

\textbf{ILE detection.} Many ILE detection studies use supervised models \citep{hu2024unified, keh2022eureka, pei2019slang, wang2022euphemism}. Their reliance on fixed training inventories is challenging because euphemisms and codewords can be replaced after becoming recognizable, a dynamic related to the \textit{euphemism treadmill} \citep{pinker2003blank, lee2022figlang, takuro2020codewords}. Weakly and semi-supervised approaches lower annotation requirements \citep{felt2020recognizing, zhang2024optimizing}, but remain vulnerable to seed sensitivity, pseudo-label noise, and error propagation. Other work uses unsupervised, self-supervised, or retrieval-based methods to discover coded terms from text \citep{ke2022unsupervised, magu2018determining, sasse2025making, takuro2020codewords, zhu2021euphemistic, zhu2021self}, , but typically requires known target or seed terms and searches within predefined domains or corpora. Recent work also explores LLM-based detection in zero- or few-shot settings \citep{fillies2024simple}. Its accuracy, however, can be sensitive to prompt formulation and to the semantic information provided about the target labels or concepts \citep{zhuo-etal-2024-prosa, gao2023benefits}.

\textbf{Existing ILE taxonomies} differ in organizing principles, methodological grounding, and scope. \citet{zhang2014appropriate} analyze Chinese entity morphs and identify phonetic substitution, character decomposition, translation, semantic reinterpretation, historical reference, and related strategies. \citet{renwick2021detection} model predefined character- and string-level transformations for normalizing obfuscated obscene language. Drawing on qualitative TikTok data, \citet{calhoun2023they}characterize linguistic self-censorship through processes involving orthographic, morphological, lexical-semantic, phonological, and prosodic manipulation, alongside multimodal practices such as gesture. \citet{fillies2024simple} organize 69 literature-derived algospeak terms into seven nonexclusive classes, while \citet{leal2024assessing} summarize nine strategies and experimentally evaluate symbol substitution and hyphenation. Related empirical taxonomies also organize intentionally noisy Korean text \citep{cho2021google}. These resources provide valuable task-specific inventories, but they were developed for different languages, platforms, and objectives and do not share an iterative procedure for testing mechanism coverage. We address this gap with a systematic taxonomy-development procedure, described next.
\section{Proposed Taxonomy}
\label{sec:taxonomy}

\subsection{Development Method}
\begin{figure}[t]
\centering
\includegraphics[width=0.85\columnwidth]{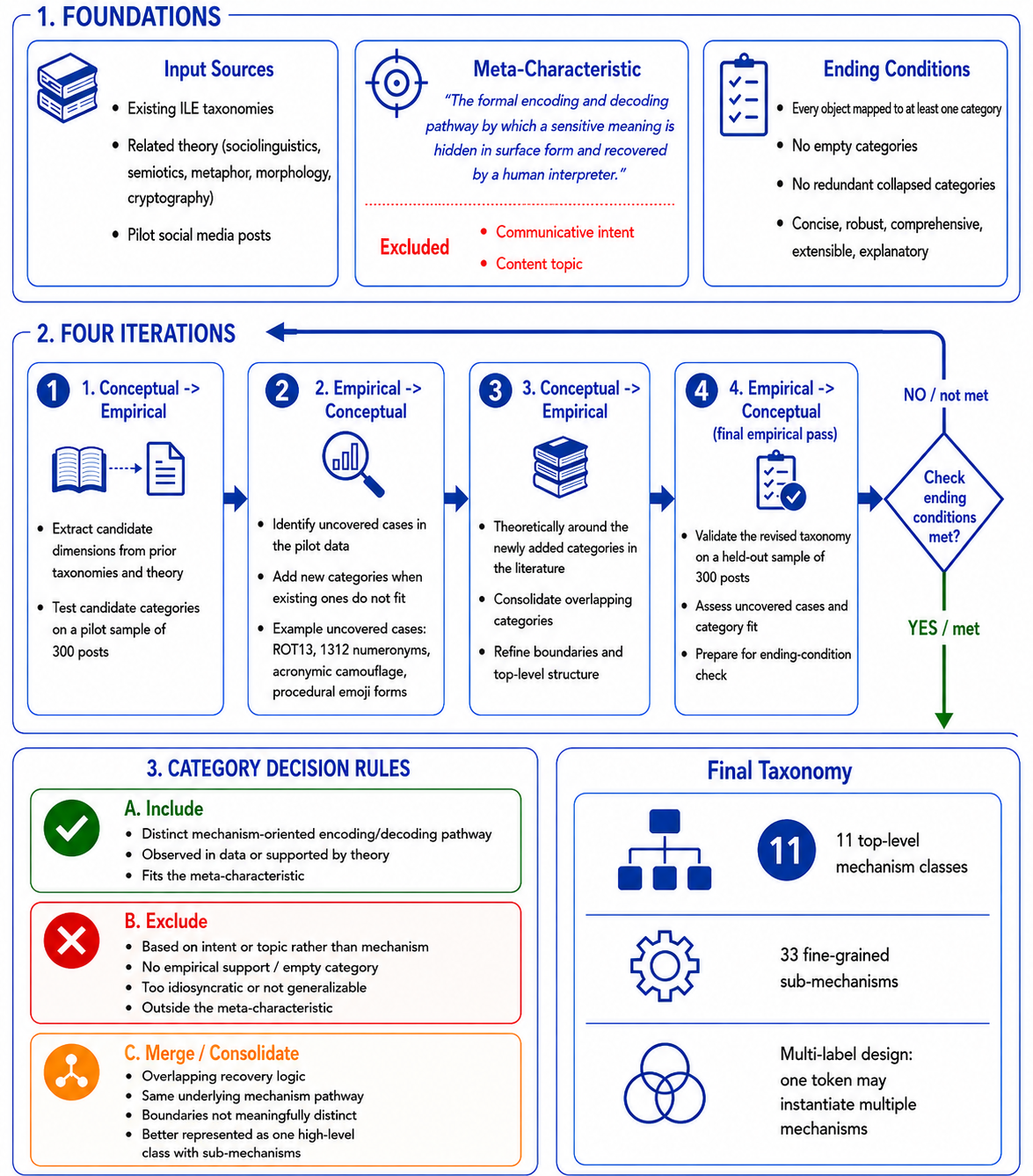}
\caption{Overview of the taxonomy development process.}
\label{fig:taxonomy_dev}
\end{figure}

We follow the iterative taxonomy development advocated by \citet{nickerson2013method}, a well-established procedure for taxonomy construction. Figure~\ref{fig:taxonomy_dev} summarizes the full development process. The method alternates between two complementary paths until a set of ending conditions is met: an \textit{empirical-to-conceptual} path that abstracts categories from observed objects, and a \textit{conceptual-to-empirical} path that derives categories from theory and then tests them on data.
\paragraph{Meta-characteristic.} In line with \citet{nickerson2013method}, we begin by defining a \textit{meta-characteristic} that serves as a logical constraint on every category in the taxonomy:
\begin{quote}
\textit{the formal encoding and decoding pathway by which a sensitive meaning is hidden in surface form and recovered by a human interpreter.}
\end{quote}
This meta-characteristic deliberately excludes communicative intent (e.g., moderation evasion, politeness, in-group signaling) and content topic (e.g., drugs, self-harm, sexuality), thereby keeping the taxonomy focused on \textit{how} meaning is concealed rather than \textit{why} or \textit{about what}.
\paragraph{Ending conditions.} Because compositional ILEs may instantiate multiple mechanisms, we adapt the objective conditions of \citet{nickerson2013method} to a nonexclusive taxonomy: all sampled objects must be covered, every class must contain at least one object, no class is added, merged, or split in the final iteration, and class definitions remain unique. We also apply their subjective conditions that the taxonomy be \textit{concise}, \textit{robust}, \textit{comprehensive}, \textit{extendible}, and \textit{explanatory}. Iterations terminate only when all conditions hold simultaneously.
\paragraph{Iterations.} The taxonomy is developed over four iterations:
\begin{enumerate}
\itemsep 0pt
\item \textbf{Conceptual-to-empirical.} We extract candidate dimensions from existing ILE taxonomies \citep{zhang2014appropriate, calhoun2023they, fillies2024simple, leal2024assessing, cho2021google, renwick2021detection} and from relevant linguistic theories pertaining to sociolinguistics, semiotics, conceptual metaphor, morphology, and cryptography. We then use a pilot sample of 300 posts to validate these dimensions.
\item \textbf{Empirical-to-conceptual.} Items not covered by any candidate category, such as ROT13 strings, numeronyms, acronymic camouflage, and procedural emoji forms, suggest new categories, with the meta-characteristic serving as the splitting criterion.
\item \textbf{Conceptual-to-empirical.} We carefully review extant literature to ground the newly added categories theoretically and consolidate those that overlap.
\item \textbf{Empirical-to-conceptual.} A final pass on a held-out 300-post sample confirms that all ending conditions are satisfied; that is, no additional new top-level category is required.
\end{enumerate}

The resulting taxonomy has 11 top-level mechanism classes and 33 fine-grained sub-mechanisms (Appendix~\ref{sec:appendix_a}). More importantly, the taxonomy does \textit{not} assume mutual exclusivity, meaning that a single encoded token can instantiate multiple mechanisms (see \S\ref{sec:discussion}).

\subsection{Top-Level Classes}
\label{sec:taxonomy_top}
Herein, we present top-level classes of the proposed taxonomy of ILE mechanisms and provide finer-grained sub-mechanisms in Appendix~\ref{sec:appendix_a}, together with definitions and examples.

\paragraph{Orthographic transformation} modifies linguistic form while often preserving human legibility despite character-level perturbations. Nonstandard orthographic variation can function as a systematic sociolinguistic resource for stylistic and indexical meaning rather than as random error \citep{androutsopoulos2011language, eckert2008variation, tagliamonte2008linguistic}. Such perturbations can disproportionately disrupt automated language processing while remaining comprehensible to human readers \citep{belinkov2017synthetic, pruthi2019combating}.
\paragraph{Phonetic substitution} alters surface form while preserving phonological similarity, allowing the intended expression to be recovered through phonological recoding and lexical access rather than orthographic form alone \citep{lukatela1994visual,wheat2010during}.
\paragraph{Formal compression} reduces expressions to compact conventional forms whose interpretation depends on learned associations and contextual expectations rather than full surface realization. This mechanism is consistent with efficiency-based accounts of language, in which predictable meanings are encoded with shorter forms to reduce production cost while preserving communicative success \citep{piantadosi2011word, zipf1949principle}.
\paragraph{Formal encoding systems} conceal meaning through defined, reversible transformations that do not require linguistic resemblance between the encoded and intended forms. Recovery proceeds by applying the corresponding decoding rule or inverse transformation, consistent with information-theoretic and cryptographic accounts of secrecy as a mapping from message space to encoded space and back  \citep{menezes2018handbook, shannon1949communication}.
\paragraph{Conventional sign reassignment} preserves the surface form of an existing sign while assigning it a new socially conventionalized meaning. Recovery depends on recognition of the learned sign–meaning association rather than rule-based transformation. Such meanings can emerge through pragmatic inference, spread, and conventionalization \citep{traugott2001regularity}, while also functioning as socially indexical resources whose values are reproduced and renegotiated in situated practice \citep{eckert2008variation}.
\paragraph{Morpho-lexical encoding} exploits productive word-formation processes to construct novel forms while retaining recognizable components; recovery proceeds by identifying the underlying lexical or morphological sources. Such formations are structured processes of lexical innovation whose realization is constrained by morphological productivity and source-word recognizability \citep{bauer2019introducing, gries2004shouldn}.
\paragraph{Referential alias encoding} conveys meaning by substituting the name of a specific identifiable entity (e.g., person, organization, platform, product) with a stabilized alternate label grounded in shared cultural or reputational knowledge. In both interactional sociolinguistics and linguistic anthropology, such meanings can be analyzed as indexical, with interpretation relying on socially recognized identities, roles, or narratives rather than direct denotation \citep{gumperz1982discourse, silverstein2003indexical}.
\paragraph{Semantic circumlocution} conveys meaning indirectly through descriptive phrasing or contextual implication rather than direct lexical expression. Recovery depends on pragmatic inference from the utterance, context, and mutually available background knowledge. This mechanism aligns with Gricean accounts of implicature, which distinguish what a speaker says from what the speaker conveys or means \citep{grice19901975}.
\paragraph{Metaphorical and metonymic encoding} exploit structured conceptual relations. Metaphor enables understanding of one conceptual domain in terms of another, whereas metonymy provides access to a target through an associated entity within the same conceptual frame, including part–whole and related relations. Cognitive-linguistic accounts treat both as conceptual mechanisms that structure ordinary thought and communication \citep{lakoff2024metaphors, radden1999metonymy}.
\paragraph{Pictorial and symbolic encoding} employs non-alphabetic signs, particularly emoji, as meaning-bearing resources. Peircean semiotics distinguishes iconic relations grounded in resemblance from symbolic relations grounded in convention \citep{atkin2010peirce, peirce1934collected}. In computer-mediated communication, emoji perform diverse pragmatic functions whose interpretation is shaped by discourse and social context, while repeated use can conventionalize and stabilize particular form–meaning associations \citep{dainas2021chapter, konrad2020sticker, weissman2023lexicon}.
\paragraph{Cross-linguistic transformation} encodes meaning by shifting across languages or scripts, including translation and transliteration. Recovery depends on knowledge of the relevant language or script correspondences needed to reconstruct the intended expression. More broadly, research on code switching and translanguaging treats multilingual and multisemiotic repertoires as flexible resources for meaning-making and social action \citep{wei2018translanguaging, gumperz1982discourse}.

\section{Dataset and Evaluations}
\label{sec:data}
\subsection{Data Collection}

\paragraph{English social-media dataset.}
We collected 2{,}000 English-language posts from two major platforms with contrasting moderation regimes: 1{,}400 TikTok video captions collected between March and May 2026, and 600 Bluesky posts collected from October 2025 to January 2026. On each platform, we used stratified sampling across two strata: posts containing at least one candidate ILE seed term or a phrase closely related to ILE usage and a topic-neutral random sample. The seed list was constructed from previous ILE studies and a pilot scan; it was intentionally broad and used only to increase the representation of plausibly coded content, rather than to determine labels. We retained posts from public-facing accounts that contained at least three tokens and were identified as English through \texttt{langid} and manual review. We removed duplicate posts, posts consisting only of URLs, hashtags, or @-mentions, and content from verified bot or automated aggregator accounts.

\paragraph{Additional cross-lingual dataset.}
We performed an additional cross-lingual evaluation, using an independently collected and annotated Chinese coded-language dataset comprising Google Maps reviews \citep{wan-etal-2026-newspaper}. We randomly selected 800 labeled reviews, including 392 coded-language-positive and 408 coded-language-negative instances. This dataset provides document- and span-level annotations and differs from our English dataset in language, platform, domain, and annotation pipeline.
\subsection{Annotation Protocol}
Two trained annotators independently labeled every post across three nested layers: \textit{document-level} ILE presence (binary), \textit{span-level} markup of the minimal contiguous span, and \textit{mechanism-level} assignment from the 11 classes in \S\ref{sec:taxonomy} and 33 sub-classes in Appendix~\ref{sec:appendix_a}, with co-occurrence marked explicitly when applicable. Importantly, document and span annotation preceded mechanism assignment. Annotators first judged from context whether a surface expression conveyed an indirect or coded meaning and marked its minimal evidence span; they did not decide ILE presence by checking whether an expression fit a proposed mechanism. Mechanism labels were assigned afterward for descriptive and class-wise analyses. 

Each annotator first completed a calibration round on 150 development items not in the released dataset. We assessed inter-annotator agreement on the full dataset using Cohen's $\kappa$ \citep{cohen1960coefficient, landis1977measurement, artstein2017inter}. The agreement is high across all annotation layers: $\kappa=0.852$ for document-level ILE presence, $\kappa=0.789$ for taxonomy class assignment, and $\kappa=0.886$ for token-level span-boundary labels. Disagreements were resolved through face-to-face discussion between the annotators until a consensus was reached. After resolution, 44.8\% of items contain at least one ILE instance.
\subsection{Models, Taxonomies, and Setup}
We evaluate six prompt variants across three LLMs. The variants comprise four prevalent taxonomies \citep{calhoun2023they, fillies2024simple, leal2024assessing, zhang2014appropriate}, the proposed taxonomy, and a \textit{no-taxonomy} baseline. The models are  \texttt{gpt-5.4}, \texttt{claude-sonnet-4-6} , and \texttt{deepseek-v4-flash}. All models are prompted in a few-shot setting with the same four-example shot set; only the embedded taxonomy section varies across prompt variants (Appendix~\ref{sec:appendix_b}).
We apply the same three LLMs and six prompt variants on the Chinese coded-language data without adapting the taxonomies to Chinese.

To contextualize LLM performance, we add four non-LLM baselines, including two supervised methods (TF-IDF + Logistic Regression and character n-gram + Linear SVM) and two unsupervised methods (Word2Vec cosine-to-seed and embedding-graph centrality). For supervised baselines, we use keyword-grouped 5-fold cross-validation (StratifiedGroupKFold), so examples sharing the same source keyword do not appear in both training and testing folds.

\begin{table*}[t]
\centering
\footnotesize
\setlength{\tabcolsep}{2.8pt}
\caption{Document- and span-level performance across LLMs and prompt variants, alongside supervised and unsupervised NLP baselines. Span-level metrics use one-to-one soft matching with $\tau=0.5$. The best per-model value in each column is \textbf{bold}.}
\label{tab:main}
\begin{tabular}{llcccc|ccc}
\toprule
\textbf{Model} & \textbf{Variant} & \multicolumn{4}{c|}{\textbf{Document Level}} & \multicolumn{3}{c}{\textbf{Span Level}} \\
 & & Accuracy & Precision & Recall & F1 & Precision & Recall & F1 \\
\midrule
\multirow{6}{*}{GPT-5.4}
 & No Taxonomy            & 0.785 & 0.811 & 0.768 & 0.771 & \textbf{0.791} & 0.483 & 0.599 \\
 & Zhang et al.           & 0.766 & 0.790 & 0.749 & 0.752 & 0.764 & 0.453 & 0.569 \\
 & Calhoun \& Fawcett     & 0.769 & 0.797 & 0.751 & 0.753 & 0.787 & 0.470 & 0.589 \\
 & Fillies \& Paschke     & 0.794 & 0.817 & 0.778 & 0.782 & 0.740 & 0.535 & 0.621 \\
 & Leal-Arenas \& Corizzo & 0.805 & 0.823 & 0.792 & 0.796 & 0.764 & 0.551 & 0.640 \\
 & Proposed Taxonomy      & \textbf{0.843} & \textbf{0.846} & \textbf{0.836} & \textbf{0.839} & 0.707 & \textbf{0.623} & \textbf{0.662} \\
\midrule
\multirow{6}{*}{Sonnet 4.6}
 & No Taxonomy            & 0.822 & \textbf{0.830} & 0.812 & 0.816 & 0.746 & 0.618 & 0.676 \\
 & Zhang et al.           & 0.788 & 0.808 & 0.773 & 0.776 & 0.771 & 0.533 & 0.630 \\
 & Calhoun \& Fawcett     & 0.806 & 0.817 & 0.794 & 0.798 & \textbf{0.778} & 0.564 & 0.654 \\
 & Fillies \& Paschke     & 0.817 & 0.821 & 0.809 & 0.812 & 0.728 & 0.624 & 0.672 \\
 & Leal-Arenas \& Corizzo & 0.823 & 0.828 & 0.814 & 0.817 & 0.771 & 0.619 & 0.686 \\
 & Proposed Taxonomy      & \textbf{0.832} & 0.831 & \textbf{0.829} & \textbf{0.830} & 0.720 & \textbf{0.660} & \textbf{0.688} \\
\midrule
\multirow{6}{*}{DeepSeek V4}
 & No Taxonomy            & 0.798 & 0.809 & 0.786 & 0.790 & 0.683 & 0.534 & 0.599 \\
 & Zhang et al.           & 0.777 & 0.799 & 0.760 & 0.764 & 0.723 & 0.448 & 0.553 \\
 & Calhoun \& Fawcett     & 0.791 & 0.806 & 0.778 & 0.782 & \textbf{0.742} & 0.506 & 0.602 \\
 & Fillies \& Paschke     & 0.807 & 0.816 & 0.796 & 0.800 & 0.693 & 0.536 & 0.605 \\
 & Leal-Arenas \& Corizzo & 0.803 & 0.813 & 0.792 & 0.796 & 0.720 & 0.563 & 0.632 \\
 & Proposed Taxonomy      & \textbf{0.828} & \textbf{0.834} & \textbf{0.819} & \textbf{0.822} & 0.713 & \textbf{0.589} & \textbf{0.645} \\
\midrule
\multirow{2}{*}{Supervised NLP}
 & TF--IDF+Logistic Regression (Grouped)
 & 0.681 & 0.678 & \textbf{0.677} & \textbf{0.677} & -- & -- & -- \\
 & Character N-Grams+SVM (Grouped)
 & \textbf{0.684} & \textbf{0.688} & 0.668 & 0.667 & -- & -- & -- \\
\midrule
\multirow{2}{*}{Unsupervised NLP}
 & Word2Vec cosine-to-seed
 & 0.515 & 0.535 & 0.531 & 0.509 & -- & -- & -- \\
 & Embedding-graph centrality
 & \textbf{0.586} & \textbf{0.577} & \textbf{0.569} & \textbf{0.564} & -- & -- & -- \\
\bottomrule
\end{tabular}
\end{table*}

\begin{table*}[t]
\centering
\footnotesize
\setlength{\tabcolsep}{4pt}
\caption{Document- and span-level performance on the Chinese coded-language dataset. Document-level P, R, and F1 are macro-averaged. Span metrics use soft matching at $\tau=0.5$. The best value in each model--metric block is shown in \textbf{bold}.}
\label{tab:codedlang}
\begin{tabular}{llcccc|ccc}
\toprule
\textbf{Model} & \textbf{Variant} & \multicolumn{4}{c|}{\textbf{Document Level}} & \multicolumn{3}{c}{\textbf{Span Level}} \\
 & & Accuracy & Precision & Recall & F1 & Precision & Recall & F1 \\
\midrule
\multirow{6}{*}{GPT-5.4} & No Taxonomy & 0.583 & 0.716 & 0.574 & 0.496 & 0.398 & 0.121 & 0.185 \\
 & Zhang et al. & 0.634 & 0.729 & 0.627 & 0.585 & 0.472 & 0.227 & 0.306 \\
 & Calhoun \& Fawcett & 0.646 & 0.761 & 0.639 & 0.596 & \textbf{0.560} & 0.288 & 0.380 \\
 & Fillies \& Paschke & 0.664 & 0.752 & 0.658 & 0.626 & 0.450 & 0.298 & 0.359 \\
 & Leal-Arenas \& Corizzo & 0.651 & 0.710 & 0.646 & 0.620 & 0.365 & 0.308 & 0.334 \\
 & Proposed Taxonomy & \textbf{0.700} & \textbf{0.785} & \textbf{0.694} & \textbf{0.672} & 0.537 & \textbf{0.360} & \textbf{0.431} \\
\midrule
\multirow{6}{*}{Sonnet 4.6} & No Taxonomy & 0.719 & 0.749 & 0.715 & 0.708 & 0.395 & 0.357 & 0.375 \\
 & Zhang et al. & 0.726 & 0.739 & 0.724 & 0.721 & 0.467 & 0.475 & 0.471 \\
 & Calhoun \& Fawcett & 0.769 & 0.799 & 0.765 & 0.761 & \textbf{0.545} & 0.475 & 0.508 \\
 & Fillies \& Paschke & 0.767 & 0.768 & 0.767 & 0.767 & 0.434 & 0.596 & 0.503 \\
 & Leal-Arenas \& Corizzo & 0.745 & 0.751 & 0.743 & 0.743 & 0.382 & 0.539 & 0.447 \\
 & Proposed Taxonomy & \textbf{0.807} & \textbf{0.808} & \textbf{0.807} & \textbf{0.807} & 0.455 & \textbf{0.616} & \textbf{0.523} \\
\midrule
\multirow{6}{*}{DeepSeek V4} & No Taxonomy & 0.549 & 0.647 & 0.540 & 0.443 & 0.451 & 0.091 & 0.152 \\
 & Zhang et al. & 0.615 & 0.731 & 0.608 & 0.552 & \textbf{0.611} & 0.217 & 0.320 \\
 & Calhoun \& Fawcett & 0.605 & 0.739 & 0.597 & 0.532 & 0.537 & 0.195 & 0.286 \\
 & Fillies \& Paschke & 0.619 & 0.730 & 0.612 & 0.559 & 0.552 & 0.224 & 0.319 \\
 & Leal-Arenas \& Corizzo & 0.601 & 0.684 & 0.594 & 0.542 & 0.520 & 0.227 & 0.316 \\
 & Proposed Taxonomy & \textbf{0.629} & \textbf{0.766} & \textbf{0.621} & \textbf{0.567} & 0.594 & \textbf{0.241} & \textbf{0.343} \\
\bottomrule
\end{tabular}
\end{table*}
\begin{table*}[t]
\centering
\scriptsize
\setlength{\tabcolsep}{2.5pt}
\caption{Per-class document-level recall for GPT-5.4. The highest value in each row is shown in \textbf{bold}. $\Delta_{\mathrm{pp}}$ is the recall difference, in percentage points, between the proposed taxonomy and the best alternative variant.}
\label{tab:class-breakdown}

\begin{tabular}{l c ccccccc}
\toprule
\textbf{Class} & \textbf{Support} & \textbf{Zhang et al.} & \textbf{Calhoun \& F.} &
\textbf{Fillies \& P.} & \textbf{Leal-A. \& C.} & \textbf{NoTax} &
\textbf{Proposed} & $\boldsymbol{\Delta_{\mathrm{pp}}}$ \\
\midrule

C1--Orthographic Transformation
& 151 & 0.795 & 0.894 & 0.881 & 0.921 & 0.795
& \textbf{0.940} & \textbf{+1.9} \\

C2--Phonetic Substitution
& 331 & 0.637 & 0.640 & 0.689 & 0.707 & 0.647
& \textbf{0.764} & \textbf{+5.7} \\

C3--Formal Compression
& 240 & 0.533 & 0.529 & 0.629 & 0.654 & 0.592
& \textbf{0.721} & \textbf{+6.7} \\

C4--Formal Encoding Systems
& 63 & 0.556 & 0.317 & 0.429 & 0.508 & 0.587
& \textbf{0.952} & \textbf{+36.5} \\

C5--Conventional Sign Reassignment
& 128 & 0.539 & 0.531 & 0.609 & 0.602 & 0.562
& \textbf{0.711} & \textbf{+10.2} \\

C6--Morpho-Lexical Encoding
& 56 & 0.661 & 0.500 & 0.554 & 0.589 & 0.518
& \textbf{0.821} & \textbf{+16.0} \\

C7--Referential Alias Encoding
& 29 & 0.828 & 0.586 & 0.655 & 0.690 & 0.621
& \textbf{1.000} & \textbf{+17.2} \\

C8--Semantic Circumlocution
& 135 & 0.570 & 0.607 & 0.644 & 0.681 & 0.652
& \textbf{0.763} & \textbf{+8.2} \\

C9--Metaphorical and Metonymic Encoding
& 45 & 0.800 & 0.733 & 0.733 & 0.822 & 0.733
& \textbf{0.933} & \textbf{+11.1} \\

C10--Pictorial and Symbolic Encoding
& 72 & 0.736 & 0.847 & 0.833 & 0.875 & 0.833
& \textbf{0.931} & \textbf{+5.6} \\

C11--Cross-Linguistic Transformation
& 10 & 0.400 & 0.400 & 0.600 & 0.500 & 0.400
& \textbf{0.800} & \textbf{+20.0} \\

\bottomrule
\end{tabular}
\end{table*}

\begin{table}[t]
\centering
\footnotesize
\setlength{\tabcolsep}{3pt}
\caption{Class-drop ablation analysis results for GPT-5.4. $\Delta$Acc and $\Delta$F1 indicate absolute differences from the full proposed taxonomy in percentage points (pp).}
\label{tab:abl}
\begin{adjustbox}{width=\columnwidth,center}
\begin{tabular}{lcccc}
\toprule
\textbf{Ablation} & \textbf{Accuracy} & \textbf{Macro-F1} &
\textbf{$\Delta$Acc} & \textbf{$\Delta$F1} \\
\midrule
Full proposed taxonomy
& 0.843 & 0.839 & 0.0 & 0.0 \\
\midrule
Drop C1 (Orthographic Transformation)
& 0.826 & 0.821 & -1.7 & -1.8 \\

Drop C2 (Phonetic Substitution)
& 0.822 & 0.815 & -2.1 & -2.4 \\

Drop C3 (Formal Compression)
& 0.824 & 0.817 & -1.9 & -2.2 \\

Drop C4 (Formal Encoding Systems)
& 0.814 & 0.808 & -2.9 & -3.1 \\

Drop C5 (Conventional Sign Reassignment)
& 0.826 & 0.820 & -1.7 & -1.9 \\

Drop C6 (Morpho-Lexical Encoding)
& 0.826 & 0.819 & -1.7 & -2.0 \\

Drop C7 (Referential Alias Encoding)
& 0.829 & 0.824 & -1.4 & -1.5 \\

Drop C8 (Semantic Circumlocution)
& 0.825 & 0.820 & -1.8 & -1.9 \\

Drop C9 (Metaphorical and Metonymic Encoding)
& 0.823 & 0.817 & -2.0 & -2.2 \\

Drop C10 (Pictorial and Symbolic Encoding)
& 0.822 & 0.816 & -2.1 & -2.3 \\

Drop C11 (Cross-Linguistic Transformation)
& 0.819 & 0.813 & -2.4 & -2.6 \\
\bottomrule
\end{tabular}
\end{adjustbox}
\end{table}

\subsection{Evaluation Metrics}

We evaluate ILE detection at two levels. At the document level, we treat the task as binary (ILE present versus absent) and report accuracy, macro-precision, macro-recall, and macro-F1. At the span level, we report precision, recall, and F1 using one-to-one soft matching between gold and predicted evidence spans. After Unicode, whitespace, and case normalization, a pair is considered a match when its similarity meets a threshold $\tau$, where similarity is the maximum of token-overlap F1 and character-overlap F1. Pairs are greedily matched in descending order of similarity, with each span matched at most once.

To test statistical significance, we use paired bootstrap resampling with 10{,}000 iterations to compute 95\% confidence intervals and two-sided $p$-values for pairwise comparisons between the model guided by the proposed taxonomy and each alternative. For document-level accuracy, we apply McNemar's test on paired predictions, which tests the null hypothesis that two classifiers do not differ systematically on the same instances \citep{McNemar_1947}. We further analyze \emph{compositional} ILEs, defined as evidence spans annotated with two or more mechanism classes, to assess whether the proposed taxonomy better captures these harder, multi-mechanism encodings. Detailed results are presented in Appendix~\ref{sec:appendix_c}, including the per-class breakdown, platform-specific results, soft matching threshold-sensitivity analysis, significance tests, and the compositional ILE analysis.

\subsection{Results}
Table~\ref{tab:main} presents document- and span-level performance across all models on the full dataset. For GPT-5.4, the proposed taxonomy achieves the best document-level performance (Acc = 0.843, macro-F1 = 0.839), exceeding the strongest competing variant (Leal-Arenas \& Corizzo) by +3.8 accuracy and +4.3 macro-F1 points. The same improvement direction holds for Sonnet~4.6 (+0.9 accuracy/+1.3 macro-F1 points) and DeepSeek~V4 (+2.1/+2.2 points), though the magnitude of gain varies by model family. At the span level, using the default soft-matching threshold $\tau=0.5$, the proposed taxonomy also attains the highest F1 for all three LLMs, showing gains of +2.2, +0.2, and +1.3 F1 points over the best-performing benchmark for GPT-5.4, Sonnet~4.6, and DeepSeek~V4, respectively. All LLM variants outperform the supervised (Acc $\leq$ 0.684) and unsupervised (Acc $\leq$ 0.586) NLP baselines by a wide margin.

The span-level improvement reflects a different precision–-recall operating point rather than an across-the-board increment in every performance metric.  Relative to the highest-precision variants, the proposed taxonomy increases span recall to 0.623, 0.660, and 0.589 for GPT-5.4, Sonnet~4.6, and DeepSeek~V4, while precision is lower (0.707, 0.720, and 0.713). This recall-oriented shift may be useful in an early-stage detector where missed ILEs cannot reach downstream review, but it also creates more false positives and should not be treated as an automatic moderation rule.

\subsection{Cross-Lingual Evaluation}
\label{sec:crosslingual}

Table~\ref{tab:codedlang} reports the cross-lingual evaluation results. At the document level, the proposed taxonomy achieves the highest accuracy, macro-precision, macro-recall, and macro-F1 across all three LLMs. Compared with the best-performing benchmark for each model, its accuracy and macro-F1 gains are +3.6 and +4.6 points for GPT-5.4, +3.8 and +4.0 points for Sonnet~4.6, and +1.0 and +0.8 points for DeepSeek~V4, respectively. We observe consistency in improvement direction, though the magnitude varies across models.

At the span level, the proposed taxonomy also achieves the highest F1 for all three LLMs, with $\tau=0.5$. It exceeds the best-performing benchmark by +5.1 points for GPT-5.4, +1.5 points for Sonnet~4.6, and +2.3 points for DeepSeek~V4. Because this dataset was collected and annotated independently, these results provide evidence suggesting that mechanism-oriented guidance can transfer to an independently annotated, cross-lingual setting.

\subsection{Class-Level Breakdown}
\label{sec:class_breakdown}

Although the aggregate metrics show that the proposed taxonomy achieves the strongest overall performance, they do not reveal which ILE mechanisms are detected more successfully. We thus compute per-class recall for each prompt variant using GPT-5.4. For each mechanism class, recall indicates how often a prompt variant detects gold documents or spans containing that mechanism. We present document-level results in Table~\ref{tab:class-breakdown} and span-level results in Appendix~\ref{sec:appendix_c}.

The proposed taxonomy achieves the highest recall across all mechanism classes. The largest difference is observed for formal encoding systems (C4; +36.5 percentage points). Excluding C11, which has only 10 instances, the next largest differences occur for referential alias encoding (C7; +17.2 points) and morpho-lexical encoding (C6; +16.0 points). These results suggest that the broader mechanism coverage of the proposed taxonomy helps GPT-5.4 detect forms of ILE that are less explicitly represented in competing taxonomies. The classes used for this breakdown analysis are defined by the proposed taxonomy, so the results should be interpreted as a diagnostic analysis of mechanism coverage rather than a direct comparison of class-label predictions. 

\subsection{Ablation Study}
\label{sec:ablation}

To assess the contribution of each class, we performed a class-drop ablation in which we removed one category at a time from the proposed taxonomy and re-evaluated GPT-5.4 using the resulting reduced prompt. Table~\ref{tab:abl} reports the absolute changes in accuracy and macro-F1, measured in percentage points, relative to the full taxonomy.

Removing any single class decreases performance, with accuracy declining by 1.4 to 2.9 percentage points and macro-F1 by 1.5 to 3.1 points. The largest degradation follows from removing formal encoding systems (C4), which reduces accuracy by 2.9 points and macro-F1 by 3.1 points. No ablated variant matches or exceeds the full taxonomy in this prompt-based setting, suggesting that each category contributes useful guidance to ILE detection.

\section{Discussion}
\label{sec:discussion}

\paragraph{Comprehensiveness makes a taxonomy useful for detection.}

The aggregate results reveal an informative pattern. For GPT-5.4, the best competing taxonomy \citep{leal2024assessing} attains a document-level macro-F1 of 0.796, while two benchmark taxonomies \citep{calhoun2023they, zhang2014appropriate} underperform the no-taxonomy prompt. More generally, taxonomy-guided variants do not consistently outperform the no-taxonomy prompt across the LLMs. That is, taxonomy guidance alone is not necessarily beneficial; its effectiveness appears to partially depend on how well the taxonomy covers the relevant encoding mechanisms. The GPT-5.4 class-level breakdown analysis supports this explanation. Alternative variants remain relatively effective for orthographic transformation, where the proposed taxonomy outperforms by only 1.9 percentage points, with larger improvements for formal encoding systems (+36.5 points), referential alias encoding (+17.2 points), and morpho-lexical encoding (+16.0 points). These diagnostic classes are defined by the proposed taxonomy, so the class-level results should not be interpreted as a taxonomy-neutral ranking. Rather, they identify the types of gold ILE evidence for which the prompt variants vary most strongly. Taken together with the class-agnostic aggregate metrics and the independently annotated Chinese coded-language evaluation, the results provide evidence that broader mechanism coverage contributes to effective prompt-guided detection. Our systematic use of a common meta-characteristic and explicit ending conditions is intended to reduce coverage gaps, though its advantage requires validation across different datasets and settings.

\paragraph{Sub-mechanism granularity matters.}
The proposed taxonomy extends existing schemes by distinguishing encoding patterns that most schemes consider as homogeneous. Its higher recall across all eleven classes (Table~\ref{tab:class-breakdown}), including classes represented broadly in prior taxonomies, reveals that finer-grained guidance may complement broader coverage. For example, pictorial and symbolic encoding distinguishes pictographic character substitution, in which a symbol replaces a character, from procedural pictographic encoding, where a symbol transforms a neighboring token. In \texttt{deew}\,\raisebox{-0.15em}{\includegraphics[height=1em]{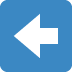}}, the left-arrow emoji functions as a reversal operator, indicating that the adjacent string should be read backward to recover \textit{weed}. Similarly, conventional sign reassignment distinguishes acronymic camouflage, in which a benign phrase such as ``Keep Yourself Safe'' represents a moderated expression through shared initials. These subclasses make the relevant recovery operations more explicit and thereby offer more precise guidance for ILE detection.
\paragraph{Comprehensive coverage makes compositional encoding tractable.}
Of the 1{,}737 ILE instances in our dataset, 257 (14.8\%) have multiple mechanisms in a single expression. For example, \texttt{161} represents AFA (``Anti-Fascist Action'') through both alphabet-position encoding and initialism. These multi-mechanism cases suggest that emerging ILEs can recombine existing mechanisms, rather than only through entirely new ones. A broad mechanism inventory can describe such hybrid expressions as compositions of known operations instead of creating a separate category for every combination.

\paragraph{Implications for content moderation.}
The proposed taxonomy enhances detection by attaining higher recall values, perhaps at the cost of precision. This tradeoff may be useful when detection is performed in an initial screening stage but cannot justify automatic enforcement. The same mechanisms can enable marginalized individuals to discuss sensitive topics while also allowing harmful actors to evade moderation. An encoding mechanism essentially identifies how meaning is concealed and recovered, without revealing why it is communicated or whether it is harmful. This separation may make the taxonomy useful as an upstream representation in task-specific systems for coded hate speech, harassment, or other policy-sensitive content. In such applications, mechanism-based detection could indicate how a message is obscured, while separate models or human reviewers assess its intent, target, context, and potential harm. These downstream applications remain to be empirically evaluated. That is, an appropriate precision–-recall balance remains application-specific.

\paragraph{An open methodological question.}
ILE evolves through both recombinations of known mechanisms and the emergence of genuinely new ones. While broad coverage can accommodate the former, the latter may require the taxonomy itself to be revised. The development procedure of \citet{nickerson2013method} specifies iterative construction and stopping conditions but not a post-deployment update lifecycle for newly emerging objects. Developing transparent criteria for identifying, validating, and incorporating new mechanisms is therefore an important direction for research on fast-moving sociolinguistic phenomena.

\section{Conclusion}
We propose a comprehensive, mechanism-oriented taxonomy of ILE and examines its efficacy in comparison with four existing taxonomies and a no-taxonomy baseline across three LLMs. With 2{,}000 manually annotated TikTok and Bluesky posts, we demonstrate that it achieves the best aggregate document- and span-level performance, though the exact gains vary across models and are generally driven by higher recall. Platform-specific and threshold-sensitivity analyses suggest that the overall pattern persists across settings, while an additional evaluation that uses 800 independently annotated Chinese coded-language reviews provides further evidence of transfer beyond the original dataset and annotation process. The empirical results underscore the importance of a comprehensive, mechanism-oriented taxonomy as a stable scaffold for detecting emerging coded language and a useful input to content moderation.

\section*{Limitations}
The primary 2{,}000-post dataset is English-only and covers two platforms and collection windows. The additional Chinese coded language data allow a cross-lingual, cross-domain evaluation yet constitute a single (external) sample that cannot establish broad linguistic, platform, temporal, or application robustness. Additional languages, independently collected datasets, and temporally separated samples are needed for robustness and generalizability. In a related sense, the reported analyses are text-only and do not cover ILE expressed through images, video, or audio.

Although aggregate document and span labels are assigned before mechanism labels and do not need to comply with our class names, mechanism-level gold labels follow the proposed taxonomy. As a result, class-wise gains—--especially for classes absent from existing schemes--—are coverage diagnostics and may benefit from definitional alignment. We do not re-annotate a taxonomy-neutral subset, restrict comparison to a formally mapped shared-class inventory, or train annotators to apply every alternative taxonomy. Thus, controlled annotation-scheme studies are needed to compare coverage, ambiguity, annotation time, and inter-annotator agreement under separately trained annotators. 

Moreover, the proposed taxonomy is examined for prompt guidance in the context of LLMs. Performance differs across models and prompts and is often associated with a trade-off between recall and precision in span-level analyses; hence, a desirable operating point must be established and confirmed for each downstream use case rather than inferred from aggregate F1 alone.

\section*{Ethical Considerations}
We use public-facing posts from TikTok and Bluesky and make no attempt to deanonymize users or infer private attributes. These publicly released data are de-identified and exclude account identifiers and other identifying metadata; TikTok-derived text is not redistributed due to the applicable data-sharing agreement. This study qualifies as IRB-exempt because it relies on public-facing, de-identified data and involves no intervention or interaction with humans. The taxonomy classifies how meaning is encoded rather than who produced it. The intended use of this work is to support content moderation and research by surfacing coded content that may evade lexicon-based filters. However, an encoding mechanism does not by itself indicate harmful intent. Mechanism-based detection should
therefore support, rather than replace, contextual judgment, policy-relevant evidence, and, where appropriate, human review. This is especially important because coded language may be used both for harmful evasion and for legitimate self-protection by vulnerable users.
The work also presents a dual-use risk: detailed decodings can help detection research but, at the same time, may reduce the effort required to reproduce or extend evasion strategies. We mitigate this risk by using primarily well-known examples, abstracting the most operationally sensitive decodings, excluding identifying metadata, and framing released artifacts around detection and evaluation rather than code generation.

\bibliography{custom-revised-citation-audited}

\appendix

\section{Extended Taxonomy Table}
\label{sec:appendix_a}
\begin{figure*}[p]
  \centering
  \includegraphics[
    width=0.8\textwidth,
    height=0.65\textheight,
    keepaspectratio
  ]{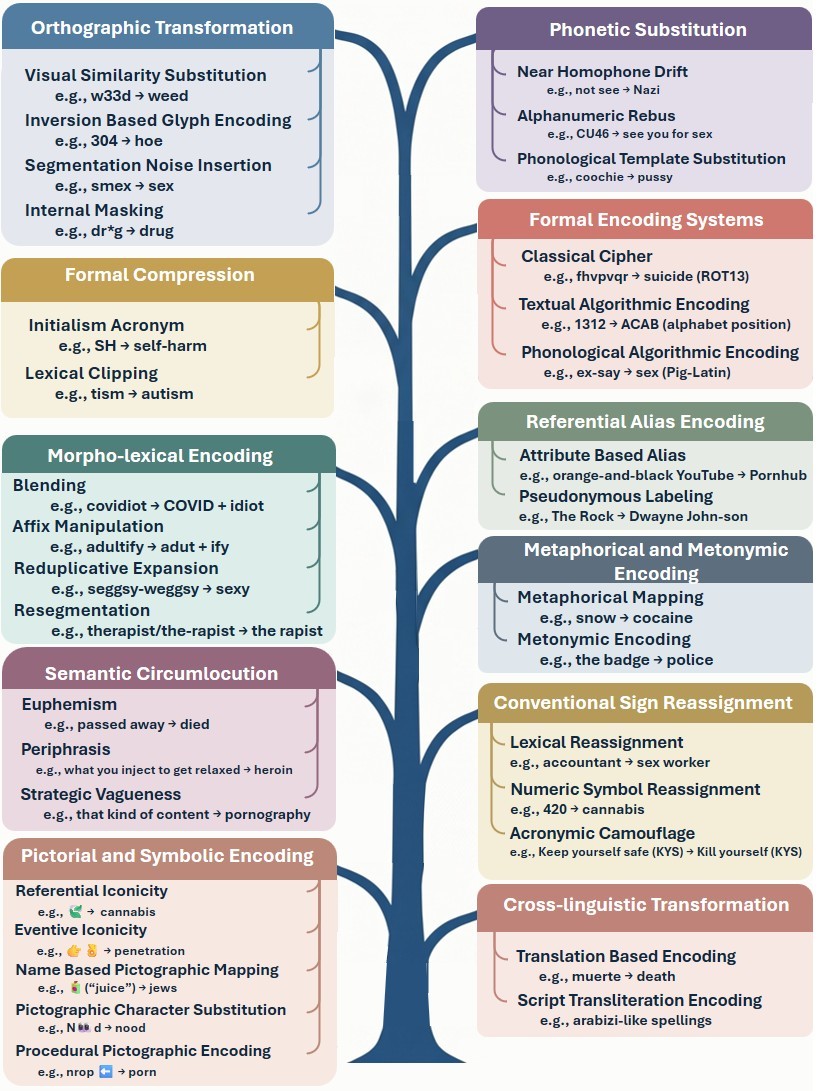}
  \caption{Overview of the proposed taxonomy: the eleven top-level mechanism classes and their corresponding sub-mechanisms.}
  \label{fig:taxonomy_viz}
\end{figure*}

Table~\ref{tab:encoding_taxonomy} presents the complete taxonomy in sub-mechanism
granularity, including operational definitions and representative examples.
Figure~\ref{fig:taxonomy_viz} provides a visual overview of the taxonomy, showing the
top-level mechanism classes and their corresponding sub-mechanisms.

\clearpage
\onecolumn

\renewcommand{\arraystretch}{1.15}
\small
\setlength{\tabcolsep}{4pt}
\setlength\LTleft{\fill}
\setlength\LTright{\fill}
\begin{longtable}{
  p{0.21\textwidth}
  p{0.50\textwidth}
  p{0.21\textwidth}
}

\caption{Sub-mechanisms of the proposed taxonomy.}
\label{tab:encoding_taxonomy} \\

\toprule
\textbf{Sub-mechanism} & \textbf{Definition} & \textbf{Examples} \\
\midrule
\endfirsthead

\toprule
\textbf{Sub-mechanism} & \textbf{Definition} & \textbf{Examples} \\
\midrule
\endhead

\midrule
\multicolumn{3}{r}{\emph{Continued on next page}} \\
\endfoot

\bottomrule
\endlastfoot

\multicolumn{3}{l}{\textbf{1. Orthographic Transformation}} \\

Visual Similarity Substitution & A target expression is altered by replacing one or more letters with visually similar characters, numbers, or symbols, while preserving the word's ordinary reading orientation. & \parbox[t]{\linewidth}{w33d $\rightarrow$ weed} \\
Inversion Based Glyph Encoding & A target expression is encoded using numbers or glyphs whose meaning becomes recoverable when the string is visually inverted, rotated, or read through a calculator-style display convention. & \parbox[t]{\linewidth}{304 $\rightarrow$ hoe} \\
Segmentation Noise Insertion & A target expression is fragmented or visually disrupted by inserting spaces, punctuation, repeated letters, or extra characters between or within letters, while preserving enough of the original sequence for recognition. & \parbox[t]{\linewidth}{k.i.l.l $\rightarrow$ kill\\smex $\rightarrow$ sex} \\
Internal Masking & One or more internal letters of a target expression are replaced or obscured with non-alphabetic symbols, while leaving enough surrounding structure for the word to remain recognizable. & \parbox[t]{\linewidth}{dr*g $\rightarrow$ drug} \\

\multicolumn{3}{l}{\textbf{2. Phonetic Substitution}} \\

Near Homophone Drift & A lexical substitute approximates the sound of a target expression while altering its spelling or lexical form. & \parbox[t]{\linewidth}{seggs $\rightarrow$ sex\\not see $\rightarrow$ Nazi} \\

Alphanumeric Rebus & Letters or numbers are used for their phonetic value to represent sounds, syllables, or words in the target expression. & \parbox[t]{\linewidth}{CU46 $\rightarrow$ see you for sex} \\

Phonological Template Substitution & A target expression is replaced by another form that preserves aspects of its phonological shape, such as syllable count, rhythm, stress pattern, or sound contour, without being a direct near-homophone. & \parbox[t]{\linewidth}{coochie $\rightarrow$ pussy} \\

\multicolumn{3}{l}{\textbf{3. Formal Compression}} \\

Initialism Acronym & A multiword target expression is compressed by retaining the initial letters, or a pronounceable subset, of its component words. & \parbox[t]{\linewidth}{SH $\rightarrow$ self-harm\\DV $\rightarrow$ Domestic Violence} \\
Lexical Clipping & A single lexical item is shortened by removing one or more segments while preserving a recognizable fragment that stands for the full form. & \parbox[t]{\linewidth}{tism $\rightarrow$ autism} \\

\multicolumn{3}{l}{\textbf{4. Formal Encoding Systems}} \\

Classical Cipher & A target expression is transformed using an established cipher or cryptographic rule, producing an encoded form that requires applying the corresponding inverse procedure to decode. & \parbox[t]{\linewidth}{fhvpvqr $\rightarrow$ suicide (ROT13)} \\
Textual Algorithmic Encoding & A target expression is encoded through an explicit rule applied to letters, numbers, or character positions, such as alphabet-index mapping, character insertion/deletion, reversal, or positional reading. & \parbox[t]{\linewidth}{1312 $\rightarrow$ ACAB (each number refers to the position of a letter in alphabetic system)\\asielx $\rightarrow$ sex (read every second character)} \\
Phonological Algorithmic Encoding & A target expression is transformed through an explicit, reversible rule applied to its sound or syllable structure, so decoding requires knowledge of the phonological transformation rule. & \parbox[t]{\linewidth}{ex-say $\rightarrow$ sex (Pig-Latin)\\puborn $\rightarrow$ porn (Ubbi Dubbi)} \\

\multicolumn{3}{l}{\textbf{5. Conventional Sign Reassignment}} \\

Lexical Reassignment & An ordinary lexical item acquires a stable coded meaning by convention, without relying on visual, phonetic, or algorithmic transformation. & \parbox[t]{\linewidth}{accountant $\rightarrow$ sex worker} \\
Conventional Numeric Symbol Reassignment & A number or symbol is assigned a stable coded meaning through social convention rather than a deterministic decoding rule. & \parbox[t]{\linewidth}{420 $\rightarrow$ cannabis\\9 $\rightarrow$ parent watching} \\
Acronymic Camouflage & A benign phrase or lexical item substitutes for a sensitive expression because both share the same acronym or initial-letter sequence. & \parbox[t]{\linewidth}{keep yourself safe (KYS) $\rightarrow$ kill yourself (KYS)} \\

\multicolumn{3}{l}{\textbf{6. Morpho-Lexical Encoding}} \\

Blending & A coded expression is formed by merging parts of two or more lexical items into a single new form, so that the source words remain partially recoverable. & \parbox[t]{\linewidth}{covidiot $\rightarrow$ COVID + idiot} \\
Affix Manipulation & A target expression is modified by adding productive prefixes, suffixes, or bound morphemes, creating a derived form whose meaning remains recoverable through the base and affixal pattern. & \parbox[t]{\linewidth}{adultify $\rightarrow$ adult + ify\\Nazify $\rightarrow$ Nazi + ify} \\
Reduplicative Expansion & A target expression is modified by repeating, echoing, or partially reduplicating sounds, syllables, or word parts to create a playful, stylized, or less direct coded form. & \parbox[t]{\linewidth}{seggsy-weggsy $\rightarrow$ sexy} \\
Resegmentation & A target expression is encoded or decoded by shifting perceived word or morpheme boundaries, so that a surface form can be re-parsed into a concealed sensitive meaning. & \parbox[t]{\linewidth}{therapist/the-rapist $\rightarrow$ the rapist} \\

\multicolumn{3}{l}{\textbf{7. Referential Alias Encoding}} \\

Attribute Based Alias & An indirect label that refers to a specific entity through a salient attribute, such as appearance, color, role, function, ownership, reputation, or platform design. & \parbox[t]{\linewidth}{the bald billionaire $\rightarrow$ Jeff Bezos\\orange-and-black YouTube $\rightarrow$ Pornhub} \\
Pseudonymous Labeling & A substitute name, nickname, or moniker assigned to a specific entity, where the link depends mainly on social convention rather than a transparent attribute. & \parbox[t]{\linewidth}{The Rock $\rightarrow$ Dwayne Johnson} \\

\multicolumn{3}{l}{\textbf{8. Semantic Circumlocution}} \\

Euphemism & A socially softer, less explicit, or less stigmatized expression replaces a direct taboo or sensitive term. & \parbox[t]{\linewidth}{passed away $\rightarrow$ died\\spicy content $\rightarrow$ pornography} \\
Periphrasis & A target concept is expressed through a longer descriptive phrase that identifies it by its function, effect, or associated activity rather than naming it directly. & \parbox[t]{\linewidth}{the thing you inject to relax yourself $\rightarrow$ heroin} \\
Strategic Vagueness & A deliberately underspecified expression avoids naming a sensitive referent, relying on context for the audience to infer the intended meaning. & \parbox[t]{\linewidth}{that kind of content $\rightarrow$ pornography} \\

\multicolumn{3}{l}{\textbf{9. Metaphorical and Metonymic Encoding}} \\

Metaphorical Mapping & A target concept is encoded through a source-domain expression whose perceived qualities, appearance, function, or symbolic associations resemble the target. & \parbox[t]{\linewidth}{snow $\rightarrow$ cocaine\\taco $\rightarrow$ female genitalia} \\
Metonymic Encoding & A target concept is encoded through a closely associated element, such as an object, place, institution, symbol, container, or part that stands for the whole. & \parbox[t]{\linewidth}{the badge $\rightarrow$ police\\the crown $\rightarrow$ monarchy} \\

\multicolumn{3}{l}{\textbf{10. Pictorial and Symbolic Encoding}} \\

Referential Iconicity & A pictorial symbol or emoji encodes a target meaning because its visual appearance directly resembles, depicts, or conventionally symbolizes the referent. & \parbox[t]{\linewidth}{\includegraphics[height=1em]{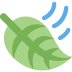} $\rightarrow$ cannabis} \\
Eventive Iconicity & A pictorial symbol or symbol combination encodes a target action, event, or practice by depicting an object, body part, gesture, or interaction associated with that action. & \parbox[t]{\linewidth}{\includegraphics[height=1em]{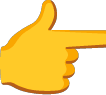}\includegraphics[height=1em]{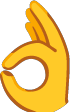} $\rightarrow$ penetration\\\includegraphics[height=1em]{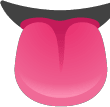} $\rightarrow$ oral sex} \\
Name Based Pictographic Mapping & A pictorial symbol encodes a target term because the symbol's name, label, or lexical reading sounds like or matches part of the target expression. & \parbox[t]{\linewidth}{\includegraphics[height=1em]{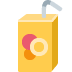}("juice") $\rightarrow$ jews} \\
Pictographic Character Substitution & An emoji or pictorial symbol replaces, masks, or interrupts one or more characters inside a word, either because its shape/name corresponds to the replaced sequence or because it functions as a visual placeholder while the surrounding letters make the target recoverable. & \parbox[t]{\linewidth}{N\includegraphics[height=1em]{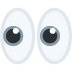}d $\rightarrow$ nood\\s\includegraphics[height=1em]{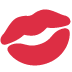}x $\rightarrow$ sex} \\
Procedural Pictographic Encoding & A pictorial symbol functions as an operator that instructs the reader to perform a decoding action, such as reversing, rotating, deleting, or substituting elements. & \parbox[t]{\linewidth}{nrop \includegraphics[height=1em]{2b05.png} $\rightarrow$ porn} \\

\multicolumn{3}{l}{\textbf{11. Cross-Linguistic Transformation}} \\

Translation Based Encoding & A target expression is replaced with its equivalent or near-equivalent in another language, allowing the sensitive meaning to be concealed from audiences or systems that primarily process the original language. & \parbox[t]{\linewidth}{muerte $\rightarrow$ death;\\chudai $\rightarrow$ sex} \\
Script Transliteration Encoding & A target expression is written in another script, romanized form, or mixed-script spelling, so the meaning remains recoverable to multilingual readers while becoming less transparent to others. & \parbox[t]{\linewidth}{arabizi-like spellings} \\

\end{longtable}

\normalsize

\clearpage
\twocolumn

\newcounter{promptctr}
\renewcommand{\thepromptctr}{\arabic{promptctr}}
\newcommand{\prompttitle}[2]{%
  \refstepcounter{promptctr}%
  \paragraph{Prompt Block \thepromptctr: #1}%
  \label{#2}%
}

\section{Prompts}
\label{sec:appendix_b}

We used two prompt templates: a no-taxonomy baseline (Prompt Block~\ref{app:dev-prompt}) and a taxonomy-aware variant (Prompt Block~\ref{app:tax-prompt}). Across taxonomy-aware conditions, only the \texttt{Categories} section of Prompt Block~\ref{app:tax-prompt} was replaced with the corresponding taxonomy's categories. All other components, including the instructions, rules, output schema, and few-shot examples, were held constant. All prompt variants, including the no-taxonomy baseline, received the same four few-shot examples. Keeping the demonstration examples constant ensures that any performance differences can be attributable to the taxonomy embedded in the prompt rather than to variation in the examples themselves. The four examples are well-established ILE cases drawn from prior work and public discussions of algospeak. We selected them mainly because every taxonomy evaluated in this study, including both the benchmark schemes and the proposed taxonomy, contains at least one mechanism capable of accounting for each example. That is, no variant is advantaged or disadvantaged by the demonstration set, since every taxonomy can, in principle, explain all four examples through its own categories. This makes the few-shot setup neutral with respect to the comparison. The complete templates are reproduced in one-column format at the end of the document.

\section{Complementary Results}
\label{sec:appendix_c}

\setlength{\dblfloatsep}{6pt plus 1pt minus 1pt}
\setlength{\dbltextfloatsep}{8pt plus 1pt minus 1pt}
\captionsetup[table]{skip=3pt}

\paragraph{Full-dataset class-specific performance.}
Table~\ref{tab:classwise_app} complements the aggregate results in Table~\ref{tab:main} by separating ILE-positive (Class~1) and ILE-negative (Class~0) document performance on the complete dataset. For GPT-5.4, the proposed taxonomy raises positive-class recall to 0.772 and F1 to 0.815, compared with 0.660 and 0.752 for the strongest benchmark, while also yielding the highest negative-class F1 (0.864). Sonnet~4.6 follows the same recall-oriented pattern: the proposed taxonomy obtains the highest positive-class recall and F1 (0.802/0.811), while its negative-class F1 (0.849) ties the no-taxonomy baseline. For DeepSeek~V4, the proposed taxonomy gives the strongest positive-class precision, recall, and F1 (0.860/0.734/0.792) and the strongest negative-class F1 (0.853). These results clarify that the aggregate gains are driven primarily by improved recovery of ILE-positive documents rather than a uniform precision increase; for example, GPT-5.4 positive-class precision is 0.863, below the best observed value of 0.879.

\paragraph{Platform-specific performance.}
Table~\ref{tab:platform-app} presents TikTok and Bluesky results side by side. Document-level precision, recall, and F1 are macro-averaged, whereas span metrics use one-to-one soft matching with a default threshold $\tau=0.5$. The proposed taxonomy achieves the highest document accuracy and macro-F1 for every model and platform. Compared with the best-performing variant, its document macro-F1 margins are 0.8/12.1 percentage points for GPT-5.4, 0.9/0.2 points for Sonnet~4.6, and 0.5/1.7 points for DeepSeek~V4 on TikTok/Bluesky, respectively. At the span level, it achieves the highest F1 in all six model--platform combination comparisons, though the TikTok margins are small: 0.6 points for GPT-5.4 and 0.1 points for both Sonnet~4.6 and DeepSeek~V4. Overall, the magnitude of improvement appears model- and platform-dependent.

\paragraph{Platform-specific class behavior.}
Table~\ref{tab:classwise-platform-app} shows that the proposed taxonomy achieves the highest positive-class recall and F1 for every model and platform, though its positive-class precision is usually lower. The notably larger GPT-5.4 gain on Bluesky is mainly driven by improved recovery of ILE-positive posts. Compared with the best alternative, the proposed taxonomy increases positive-class recall from 0.571 to 0.792, while its precision remains mostly intact (0.973 versus 0.977). This pattern is consistent with broader mechanism coverage that helps GPT-5.4 recover positive instances missed under narrower taxonomy guidance. Negative-class F1 trails the best alternative in only three cases: by 0.1 percentage points for GPT-5.4 on TikTok and by 0.3 and 0.2 points for Sonnet~4.6 on TikTok and Bluesky, respectively. Overall, the class-specific results indicate that the proposed taxonomy's advantage is mainly recall-driven.

\paragraph{Soft-matching threshold sensitivity.}
Table~\ref{tab:threshold-sensitivity-app} reports span performance at $\tau\in\{0.4,0.5,0.6\}$. The proposed taxonomy achieves the highest recall in all 18 model--platform--threshold combination comparisons; its F1 is the highest or tied in 17. The only exception is Sonnet~4.6 on Bluesky at $\tau=0.6$, where the best alternative taxonomy is higher in F1 by 0.1 percentage points (0.681 versus 0.680). In summary, the proposed taxonomy’s advantage in F1 is stable across thresholds, remaining recall-driven and occasionally lower.


\clearpage

\begin{table*}[!t]
\centering
\scriptsize
\setlength{\tabcolsep}{2pt}
\caption{Full-dataset class-wise document-level performance across models and prompt variants. Class~1 denotes ILE-positive documents and Class~0 denotes ILE-negative documents. The best value for each model block is shown in \textbf{bold}.}
\label{tab:classwise_app}
\begin{tabular*}{\textwidth}{@{\extracolsep{\fill}}llccc|ccc}
\toprule
\textbf{Model} & \textbf{Variant} & \multicolumn{3}{c|}{\textbf{Class 1 (ILE)}} & \multicolumn{3}{c}{\textbf{Class 0}} \\
 & & Precision & Recall & F1 & Precision & Recall & F1 \\
\midrule
\multirow{6}{*}{GPT-5.4}
 & No Taxonomy            & \textbf{0.879} & 0.603 & 0.715 & 0.743 & \textbf{0.933} & 0.827 \\
 & Zhang et al.           & 0.850 & 0.581 & 0.691 & 0.730 & 0.917 & 0.813 \\
 & Calhoun \& Fawcett     & 0.865 & 0.574 & 0.690 & 0.728 & 0.928 & 0.816 \\
 & Fillies \& Paschke     & \textbf{0.879} & 0.626 & 0.731 & 0.754 & 0.930 & 0.833 \\
 & Leal-Arenas \& Corizzo & 0.876 & 0.660 & 0.752 & 0.770 & 0.924 & 0.840 \\
 & Proposed Taxonomy      & 0.863 & \textbf{0.772} & \textbf{0.815} & \textbf{0.830} & 0.900 & \textbf{0.864} \\
\midrule
\multirow{6}{*}{Sonnet 4.6}
 & No Taxonomy            & 0.861 & 0.719 & 0.783 & 0.799 & 0.906 & \textbf{0.849} \\
 & Zhang et al.           & \textbf{0.865} & 0.624 & 0.725 & 0.751 & \textbf{0.921} & 0.828 \\
 & Calhoun \& Fawcett     & 0.857 & 0.681 & 0.759 & 0.778 & 0.908 & 0.838 \\
 & Fillies \& Paschke     & 0.838 & 0.733 & 0.782 & 0.803 & 0.885 & 0.842 \\
 & Leal-Arenas \& Corizzo & 0.854 & 0.729 & 0.786 & 0.803 & 0.899 & 0.848 \\
 & Proposed Taxonomy      & 0.819 & \textbf{0.802} & \textbf{0.811} & \textbf{0.842} & 0.856 & \textbf{0.849} \\
\midrule
\multirow{6}{*}{DeepSeek V4}
 & No Taxonomy            & 0.848 & 0.669 & 0.748 & 0.770 & 0.903 & 0.832 \\
 & Zhang et al.           & 0.858 & 0.602 & 0.707 & 0.740 & \textbf{0.919} & 0.820 \\
 & Calhoun \& Fawcett     & 0.852 & 0.647 & 0.736 & 0.760 & 0.909 & 0.828 \\
 & Fillies \& Paschke     & 0.849 & 0.692 & 0.763 & 0.783 & 0.900 & 0.837 \\
 & Leal-Arenas \& Corizzo & 0.849 & 0.683 & 0.757 & 0.778 & 0.901 & 0.835 \\
 & Proposed Taxonomy      & \textbf{0.860} & \textbf{0.734} & \textbf{0.792} & \textbf{0.807} & 0.903 & \textbf{0.853} \\
\midrule
\multirow{2}{*}{Supervised NLP}
 & TF--IDF+Logistic Regression (Grouped)
 & 0.648 & \textbf{0.634} & \textbf{0.641} & \textbf{0.707} & 0.720 & 0.714 \\
 & Character N-Grams/SVM (Grouped)
 & \textbf{0.700} & 0.514 & 0.593 & 0.676 & \textbf{0.821} & \textbf{0.740} \\
\midrule
\multirow{2}{*}{Unsupervised NLP}
 & Word2Vec cosine--to--seed
 & 0.472 & \textbf{0.690} & \textbf{0.560} & 0.597 & 0.372 & 0.458 \\
 & Embedding-graph centrality
 & \textbf{0.551} & 0.408 & 0.469 & \textbf{0.603} & \textbf{0.730} & \textbf{0.660} \\
\bottomrule
\end{tabular*}
\end{table*}

\begin{table*}[!t]
\centering
\scriptsize
\setlength{\tabcolsep}{1pt}
\caption{Platform-specific document- and span-level performance across LLMs and prompt variants. Document metrics are macro-averaged where applicable. Span metrics use one-to-one soft matching with $\tau=0.5$. The best value for each model--platform block is shown in \textbf{bold}.}
\label{tab:platform-app}
\begin{tabular*}{\textwidth}{@{\extracolsep{\fill}}llcccc|ccc|cccc|ccc}
\toprule
\textbf{Model} & \textbf{Variant} & \multicolumn{7}{c|}{\textbf{TikTok}} & \multicolumn{7}{c}{\textbf{Bluesky}} \\
\cmidrule(lr){3-9} \cmidrule(lr){10-16}
 & & \multicolumn{4}{c|}{\textbf{Document-level}} & \multicolumn{3}{c|}{\textbf{Span-level}} & \multicolumn{4}{c|}{\textbf{Document-level}} & \multicolumn{3}{c}{\textbf{Span-level}} \\
 & & Acc & P & R & F1 & P & R & F1 & Acc & P & R & F1 & P & R & F1 \\
\midrule
\multirow{6}{*}{GPT-5.4} & No Taxonomy & 0.799 & 0.810 & 0.774 & 0.782 & \textbf{0.776} & 0.514 & 0.619 & 0.753 & 0.816 & 0.766 & 0.746 & 0.832 & 0.416 & 0.555 \\
 & Zhang et al. & 0.779 & 0.789 & 0.754 & 0.761 & 0.743 & 0.468 & 0.574 & 0.737 & 0.792 & 0.749 & 0.729 & 0.819 & 0.421 & 0.557 \\
 & Calhoun \& Fawcett & 0.791 & 0.802 & 0.766 & 0.773 & 0.768 & 0.516 & 0.617 & 0.718 & 0.793 & 0.732 & 0.706 & \textbf{0.850} & 0.373 & 0.519 \\
 & Fillies \& Paschke & 0.810 & 0.818 & 0.789 & 0.796 & 0.716 & 0.576 & 0.639 & 0.757 & 0.820 & 0.769 & 0.749 & 0.815 & 0.448 & 0.578 \\
 & Leal-Arenas \& Corizzo & 0.823 & \textbf{0.826} & 0.806 & 0.812 & 0.743 & 0.597 & 0.662 & 0.765 & 0.822 & 0.777 & 0.759 & 0.827 & 0.454 & 0.586 \\
 & Proposed Taxonomy & \textbf{0.827} & 0.824 & \textbf{0.817} & \textbf{0.820} & 0.694 & \textbf{0.644} & \textbf{0.668} & \textbf{0.880} & \textbf{0.892} & \textbf{0.885} & \textbf{0.880} & 0.739 & \textbf{0.577} & \textbf{0.648} \\
\midrule
\multirow{6}{*}{Sonnet 4.6} & No Taxonomy & 0.831 & 0.830 & 0.820 & 0.824 & 0.713 & 0.651 & 0.681 & 0.800 & 0.835 & 0.809 & 0.797 & 0.846 & 0.548 & 0.665 \\
 & Zhang et al. & 0.804 & 0.813 & 0.781 & 0.789 & \textbf{0.758} & 0.571 & 0.652 & 0.752 & 0.797 & 0.762 & 0.746 & 0.808 & 0.452 & 0.580 \\
 & Calhoun \& Fawcett & 0.812 & 0.813 & 0.797 & 0.802 & 0.752 & 0.598 & 0.667 & 0.792 & 0.830 & 0.801 & 0.789 & 0.854 & 0.491 & 0.624 \\
 & Fillies \& Paschke & 0.828 & 0.823 & 0.821 & 0.822 & 0.699 & 0.670 & 0.684 & 0.792 & 0.822 & 0.800 & 0.789 & 0.815 & 0.527 & 0.640 \\
 & Leal-Arenas \& Corizzo & 0.824 & 0.823 & 0.811 & 0.815 & 0.737 & 0.642 & 0.686 & 0.820 & \textbf{0.837} & 0.826 & 0.819 & \textbf{0.867} & 0.570 & 0.688 \\
 & Proposed Taxonomy & \textbf{0.836} & \textbf{0.831} & \textbf{0.837} & \textbf{0.833} & 0.685 & \textbf{0.689} & \textbf{0.687} & \textbf{0.822} & 0.832 & \textbf{0.827} & \textbf{0.821} & 0.821 & \textbf{0.598} & \textbf{0.692} \\
\midrule
\multirow{6}{*}{DeepSeek V4} & No Taxonomy & 0.787 & 0.791 & 0.766 & 0.773 & 0.694 & 0.542 & 0.609 & 0.823 & 0.843 & 0.830 & 0.822 & 0.661 & 0.516 & 0.580 \\
 & Zhang et al. & 0.770 & 0.783 & 0.741 & 0.748 & 0.697 & 0.430 & 0.532 & 0.793 & 0.823 & 0.802 & 0.791 & 0.777 & 0.486 & 0.598 \\
 & Calhoun \& Fawcett & 0.786 & 0.792 & 0.764 & 0.771 & \textbf{0.717} & 0.515 & 0.600 & 0.803 & 0.830 & 0.811 & 0.802 & \textbf{0.803} & 0.487 & 0.607 \\
 & Fillies \& Paschke & 0.794 & 0.796 & 0.775 & 0.781 & 0.671 & 0.542 & 0.600 & 0.838 & 0.853 & 0.844 & 0.838 & 0.747 & 0.523 & 0.616 \\
 & Leal-Arenas \& Corizzo & 0.812 & 0.814 & 0.796 & 0.801 & 0.714 & 0.591 & 0.647 & 0.783 & 0.810 & 0.791 & 0.781 & 0.737 & 0.505 & 0.600 \\
 & Proposed Taxonomy & \textbf{0.816} & \textbf{0.816} & \textbf{0.800} & \textbf{0.806} & 0.697 & \textbf{0.606} & \textbf{0.648} & \textbf{0.855} & \textbf{0.864} & \textbf{0.860} & \textbf{0.855} & 0.756 & \textbf{0.552} & \textbf{0.638} \\
\bottomrule
\end{tabular*}
\vspace{2pt}
\begin{minipage}{\textwidth}
\scriptsize
\textit{Note:} Acc: Accuracy; P: Precision; R: Recall.
\end{minipage}
\end{table*}


\clearpage

\begin{table*}[!t]
\centering
\scriptsize
\setlength{\tabcolsep}{1.5pt}
\caption{Platform-specific class-wise document-level performance across LLMs and prompt variants. Class~1 denotes ILE-positive documents and Class~0 denotes ILE-negative documents. The best value for each model--platform block is shown in \textbf{bold}.}
\label{tab:classwise-platform-app}
\begin{tabular*}{\textwidth}{@{\extracolsep{\fill}}llccc|ccc|ccc|ccc}
\toprule
\textbf{Model} & \textbf{Variant} & \multicolumn{6}{c|}{\textbf{TikTok}} & \multicolumn{6}{c}{\textbf{Bluesky}} \\
\cmidrule(lr){3-8} \cmidrule(lr){9-14}
 & & \multicolumn{3}{c|}{\textbf{Class 1}} & \multicolumn{3}{c|}{\textbf{Class 0}} & \multicolumn{3}{c|}{\textbf{Class 1}} & \multicolumn{3}{c}{\textbf{Class 0}} \\
 & & P & R & F1 & P & R & F1 & P & R & F1 & P & R & F1 \\
\midrule
\multirow{6}{*}{GPT-5.4}
 & No Taxonomy & \textbf{0.841} & 0.632 & 0.722 & 0.779 & \textbf{0.916} & 0.842 & 0.972 & 0.549 & 0.702 & 0.660 & 0.982 & 0.790 \\
 & Zhang et al. & 0.814 & 0.604 & 0.694 & 0.764 & 0.903 & 0.827 & 0.934 & 0.539 & 0.684 & 0.650 & 0.958 & 0.774 \\
 & Calhoun \& Fawcett & 0.831 & 0.620 & 0.710 & 0.773 & 0.911 & 0.836 & 0.957 & 0.489 & 0.647 & 0.630 & 0.975 & 0.766 \\
 & Fillies \& Paschke & \textbf{0.841} & 0.667 & 0.744 & 0.795 & 0.911 & 0.849 & \textbf{0.978} & 0.552 & 0.706 & 0.663 & \textbf{0.986} & 0.793 \\
 & Leal-Arenas \& Corizzo & 0.838 & 0.708 & 0.768 & 0.814 & 0.904 & \textbf{0.857} & 0.973 & 0.571 & 0.720 & 0.671 & 0.982 & 0.798 \\
 & Proposed Taxonomy & 0.809 & \textbf{0.762} & \textbf{0.785} & \textbf{0.839} & 0.873 & 0.856 & 0.977 & \textbf{0.792} & \textbf{0.875} & \textbf{0.808} & 0.979 & \textbf{0.885} \\
\midrule
\multirow{6}{*}{Sonnet 4.6}
 & No Taxonomy & 0.822 & 0.756 & 0.788 & 0.837 & 0.884 & \textbf{0.860} & \textbf{0.958} & 0.650 & 0.774 & 0.712 & \textbf{0.968} & 0.820 \\
 & Zhang et al. & \textbf{0.838} & 0.651 & 0.733 & 0.787 & \textbf{0.911} & 0.845 & 0.929 & 0.574 & 0.710 & 0.666 & 0.951 & 0.783 \\
 & Calhoun \& Fawcett & 0.815 & 0.706 & 0.757 & 0.811 & 0.887 & 0.847 & 0.957 & 0.634 & 0.763 & 0.703 & \textbf{0.968} & 0.814 \\
 & Fillies \& Paschke & 0.800 & 0.779 & 0.789 & 0.847 & 0.862 & 0.855 & 0.936 & 0.650 & 0.767 & 0.708 & 0.951 & 0.811 \\
 & Leal-Arenas \& Corizzo & 0.819 & 0.736 & 0.775 & 0.826 & 0.886 & 0.855 & 0.927 & 0.716 & 0.808 & 0.746 & 0.936 & \textbf{0.831} \\
 & Proposed Taxonomy & 0.782 & \textbf{0.838} & \textbf{0.809} & \textbf{0.879} & 0.836 & 0.857 & 0.907 & \textbf{0.738} & \textbf{0.814} & \textbf{0.757} & 0.915 & 0.829 \\
\midrule
\multirow{6}{*}{DeepSeek V4}
 & No Taxonomy & 0.801 & 0.646 & 0.715 & 0.780 & 0.887 & 0.830 & \textbf{0.941} & 0.710 & 0.809 & 0.745 & \textbf{0.951} & 0.835 \\
 & Zhang et al. & 0.816 & 0.573 & 0.673 & 0.751 & \textbf{0.909} & 0.822 & 0.937 & 0.653 & 0.770 & 0.710 & \textbf{0.951} & 0.813 \\
 & Calhoun \& Fawcett & 0.808 & 0.634 & 0.711 & 0.776 & 0.894 & 0.831 & 0.938 & 0.672 & 0.783 & 0.721 & \textbf{0.951} & 0.820 \\
 & Fillies \& Paschke & 0.802 & 0.665 & 0.727 & 0.789 & 0.884 & 0.834 & 0.940 & 0.741 & 0.829 & 0.766 & 0.947 & 0.847 \\
 & Leal-Arenas \& Corizzo & \textbf{0.819} & 0.701 & 0.755 & 0.809 & 0.890 & 0.848 & 0.916 & 0.650 & 0.760 & 0.704 & 0.933 & 0.802 \\
 & Proposed Taxonomy & \textbf{0.819} & \textbf{0.712} & \textbf{0.762} & \textbf{0.814} & 0.889 & \textbf{0.850} & 0.939 & \textbf{0.776} & \textbf{0.850} & \textbf{0.790} & 0.943 & \textbf{0.860} \\
\bottomrule
\end{tabular*}
\vspace{2pt}
\begin{minipage}{\textwidth}
\scriptsize
\textit{Note:} P: Precision; R: Recall.
\end{minipage}
\end{table*}

\begin{table*}[!t]
\centering
\scriptsize
\setlength{\tabcolsep}{0.25pt}
\caption{Soft-span precision, recall, and F1 threshold sensitivity across TikTok and Bluesky. The best value for each model--platform--threshold--metric block is shown in \textbf{bold}.}
\label{tab:threshold-sensitivity-app}
\begin{tabular*}{\textwidth}{@{\extracolsep{\fill}}llccc|ccc|ccc|ccc|ccc|ccc}
\toprule
\textbf{Model} & \textbf{Variant} & \multicolumn{9}{c|}{\textbf{TikTok}} & \multicolumn{9}{c}{\textbf{Bluesky}} \\
\cmidrule(lr){3-11} \cmidrule(lr){12-20}
 & & \multicolumn{3}{c|}{$\boldsymbol{\tau=.4}$} & \multicolumn{3}{c|}{$\boldsymbol{\tau=.5}$} & \multicolumn{3}{c|}{$\boldsymbol{\tau=.6}$} & \multicolumn{3}{c|}{$\boldsymbol{\tau=.4}$} & \multicolumn{3}{c|}{$\boldsymbol{\tau=.5}$} & \multicolumn{3}{c}{$\boldsymbol{\tau=.6}$} \\
 & & P & R & F1 & P & R & F1 & P & R & F1 & P & R & F1 & P & R & F1 & P & R & F1 \\
\midrule
\multirow{6}{*}{GPT-5.4} & No Taxonomy & \textbf{0.786} & 0.521 & 0.627 & \textbf{0.776} & 0.514 & 0.619 & \textbf{0.761} & 0.504 & 0.606 & 0.846 & 0.423 & 0.564 & 0.832 & 0.416 & 0.555 & 0.811 & 0.405 & 0.540 \\
 & Zhang et al. & 0.752 & 0.474 & 0.581 & 0.743 & 0.468 & 0.574 & 0.732 & 0.461 & 0.566 & 0.830 & 0.427 & 0.564 & 0.819 & 0.421 & 0.557 & 0.781 & 0.402 & 0.531 \\
 & Calhoun \& Fawcett & 0.773 & 0.520 & 0.621 & 0.768 & 0.516 & 0.617 & 0.755 & 0.508 & 0.607 & \textbf{0.854} & 0.375 & 0.521 & \textbf{0.850} & 0.373 & 0.519 & \textbf{0.829} & 0.364 & 0.506 \\
 & Fillies \& Paschke & 0.722 & 0.581 & 0.643 & 0.716 & 0.576 & 0.639 & 0.703 & 0.565 & 0.627 & 0.821 & 0.452 & 0.583 & 0.815 & 0.448 & 0.578 & 0.795 & 0.438 & 0.565 \\
 & Leal-Arenas \& Corizzo & 0.749 & 0.602 & 0.668 & 0.743 & 0.597 & 0.662 & 0.729 & 0.586 & 0.650 & 0.847 & 0.464 & 0.600 & 0.827 & 0.454 & 0.586 & 0.798 & 0.438 & 0.565 \\
 & Proposed Taxonomy & 0.699 & \textbf{0.649} & \textbf{0.673} & 0.694 & \textbf{0.644} & \textbf{0.668} & 0.685 & \textbf{0.636} & \textbf{0.659} & 0.769 & \textbf{0.600} & \textbf{0.674} & 0.739 & \textbf{0.577} & \textbf{0.648} & 0.691 & \textbf{0.539} & \textbf{0.606} \\
\midrule
\multirow{6}{*}{Sonnet 4.6} & No Taxonomy & 0.727 & 0.664 & 0.694 & 0.713 & 0.651 & 0.681 & 0.696 & 0.636 & 0.665 & 0.860 & 0.557 & 0.676 & 0.846 & 0.548 & 0.665 & 0.824 & 0.534 & 0.648 \\
 & Zhang et al. & \textbf{0.767} & 0.578 & 0.659 & \textbf{0.758} & 0.571 & 0.652 & \textbf{0.749} & 0.565 & 0.644 & 0.827 & 0.463 & 0.593 & 0.808 & 0.452 & 0.580 & 0.789 & 0.441 & 0.566 \\
 & Calhoun \& Fawcett & 0.761 & 0.605 & 0.674 & 0.752 & 0.598 & 0.667 & 0.743 & 0.591 & 0.658 & 0.870 & 0.500 & 0.635 & 0.854 & 0.491 & 0.624 & 0.835 & 0.480 & 0.610 \\
 & Fillies \& Paschke & 0.709 & 0.679 & 0.694 & 0.699 & 0.670 & 0.684 & 0.686 & 0.657 & 0.671 & 0.829 & 0.536 & 0.651 & 0.815 & 0.527 & 0.640 & 0.793 & 0.513 & 0.623 \\
 & Leal-Arenas \& Corizzo & 0.748 & 0.651 & 0.696 & 0.737 & 0.642 & 0.686 & 0.723 & 0.630 & 0.673 & \textbf{0.872} & 0.573 & 0.692 & \textbf{0.867} & 0.570 & 0.688 & \textbf{0.859} & 0.564 & \textbf{0.681} \\
 & Proposed Taxonomy & 0.697 & \textbf{0.702} & \textbf{0.700} & 0.685 & \textbf{0.689} & \textbf{0.687} & 0.671 & \textbf{0.676} & \textbf{0.673} & 0.833 & \textbf{0.607} & \textbf{0.702} & 0.821 & \textbf{0.598} & \textbf{0.692} & 0.806 & \textbf{0.588} & 0.680 \\
\midrule
\multirow{6}{*}{DeepSeek V4} & No Taxonomy & 0.701 & 0.548 & 0.616 & 0.694 & 0.542 & 0.609 & 0.680 & 0.531 & 0.596 & 0.682 & 0.532 & 0.598 & 0.661 & 0.516 & 0.580 & 0.632 & 0.493 & 0.554 \\
 & Zhang et al. & 0.704 & 0.435 & 0.538 & 0.697 & 0.430 & 0.532 & 0.688 & 0.424 & 0.525 & 0.794 & 0.496 & 0.611 & 0.777 & 0.486 & 0.598 & 0.743 & 0.464 & 0.571 \\
 & Calhoun \& Fawcett & 0.720 & 0.517 & 0.602 & \textbf{0.717} & 0.515 & 0.600 & \textbf{0.704} & 0.506 & 0.589 & \textbf{0.824} & 0.500 & 0.622 & \textbf{0.803} & 0.488 & 0.607 & \textbf{0.788} & 0.479 & 0.596 \\
 & Fillies \& Paschke & 0.678 & 0.548 & 0.606 & 0.671 & 0.542 & 0.600 & 0.658 & 0.532 & 0.588 & 0.773 & 0.541 & 0.637 & 0.747 & 0.523 & 0.616 & 0.709 & 0.496 & 0.584 \\
 & Leal-Arenas \& Corizzo & \textbf{0.720} & 0.596 & 0.652 & 0.714 & 0.591 & 0.647 & 0.699 & 0.579 & 0.634 & 0.745 & 0.511 & 0.606 & 0.737 & 0.505 & 0.600 & 0.727 & 0.498 & 0.591 \\
 & Proposed Taxonomy & 0.701 & \textbf{0.610} & \textbf{0.653} & 0.697 & \textbf{0.606} & \textbf{0.648} & 0.686 & \textbf{0.597} & \textbf{0.638} & 0.795 & \textbf{0.580} & \textbf{0.671} & 0.756 & \textbf{0.552} & \textbf{0.638} & 0.716 & \textbf{0.523} & \textbf{0.605} \\
\bottomrule
\end{tabular*}
\vspace{2pt}
\begin{minipage}{\textwidth}
\scriptsize
\textit{Note:} P: Precision; R: Recall.
\end{minipage}
\end{table*}


\clearpage

\twocolumn[
\begin{minipage}{\textwidth}
\centering
\scriptsize
\setlength{\tabcolsep}{4pt}
\captionof{table}{Class-wise document performance on the Chinese coded-language data. Class~1 denotes coded-language-positive documents and Class~0 denotes coded-language-negative documents. The best value for each model block is shown in \textbf{bold}.}
\label{tab:classwise-chinese-app}
\begin{adjustbox}{max width=\textwidth}
\begin{tabular}{llccc|ccc}
\toprule
\textbf{Model} & \textbf{Variant} & \multicolumn{3}{c|}{\textbf{Class 1 (coded)}} & \multicolumn{3}{c}{\textbf{Class 0}} \\
 & & Precision & Recall & F1 & Precision & Recall & F1 \\
\midrule
\multirow{6}{*}{GPT-5.4} & No Taxonomy & 0.882 & 0.171 & 0.286 & 0.551 & \textbf{0.978} & 0.705 \\
 & Zhang et al. & 0.872 & 0.296 & 0.442 & 0.586 & 0.958 & 0.727 \\
 & Calhoun \& Fawcett & 0.929 & 0.301 & 0.455 & 0.593 & \textbf{0.978} & 0.738 \\
 & Fillies \& Paschke & 0.897 & 0.355 & 0.508 & 0.608 & 0.961 & 0.745 \\
 & Leal-Arenas \& Corizzo & 0.816 & 0.372 & 0.511 & 0.604 & 0.919 & 0.729 \\
 & Proposed Taxonomy & \textbf{0.937} & \textbf{0.416} & \textbf{0.576} & \textbf{0.634} & 0.973 & \textbf{0.768} \\
\midrule
\multirow{6}{*}{Sonnet 4.6} & No Taxonomy & 0.830 & 0.536 & 0.651 & 0.667 & 0.895 & 0.764 \\
 & Zhang et al. & 0.789 & 0.602 & 0.683 & 0.689 & 0.846 & 0.759 \\
 & Calhoun \& Fawcett & \textbf{0.891} & 0.602 & 0.718 & 0.708 & \textbf{0.929} & 0.804 \\
 & Fillies \& Paschke & 0.777 & 0.737 & 0.757 & 0.759 & 0.797 & 0.778 \\
 & Leal-Arenas \& Corizzo & 0.783 & 0.663 & 0.718 & 0.718 & 0.824 & 0.767 \\
 & Proposed Taxonomy & 0.815 & \textbf{0.786} & \textbf{0.800} & \textbf{0.801} & 0.828 & \textbf{0.814} \\
\midrule
\multirow{6}{*}{DeepSeek V4} & No Taxonomy & 0.763 & 0.115 & 0.200 & 0.532 & 0.966 & 0.686 \\
 & Zhang et al. & 0.889 & 0.245 & 0.384 & 0.572 & 0.971 & 0.720 \\
 & Calhoun \& Fawcett & 0.913 & 0.214 & 0.347 & 0.565 & 0.980 & 0.717 \\
 & Fillies \& Paschke & 0.885 & \textbf{0.255} & 0.396 & 0.575 & 0.968 & 0.721 \\
 & Leal-Arenas \& Corizzo & 0.802 & 0.247 & 0.378 & 0.566 & 0.941 & 0.707 \\
 & Proposed Taxonomy & \textbf{0.952} & \textbf{0.255} & \textbf{0.402} & \textbf{0.580} & \textbf{0.988} & \textbf{0.731} \\
\bottomrule
\end{tabular}
\end{adjustbox}
\vspace{4pt}
\end{minipage}
]

\paragraph{Cross-lingual evaluation.}
Table~\ref{tab:classwise-chinese-app} indicates that the proposed taxonomy achieves the highest coded-language-positive and negative-class F1 across all three LLMs. Its positive-class F1 improvement over the best-performing alternative is 6.5 percentage points for GPT-5.4, 4.3 points for Sonnet~4.6, and 0.6 points for DeepSeek~V4. The smaller DeepSeek improvement again suggests the magnitude of its improvement to be model-dependent.

Table~\ref{tab:threshold-sensitivity-chinese-app} presents soft-span precision, recall, and F1 under different thresholds. As shown, the proposed taxonomy achieves the highest recall and F1 for all three LLMs with $\tau=0.4$, 0.5, and 0.6, while alternative variants generally achieve higher precision. Its F1 improvement over the best-performing alternative is 5.0/5.1/4.3 percentage points for GPT-5.4, 0.7/1.5/1.0 points for Sonnet~4.6, and 1.9/2.3/1.0 points for DeepSeek~V4 across the three different thresholds, respectively. The proposed taxonomy exhibits consistent cross-lingual, span-level advantage across thresholds, but its magnitude of improvement differs with models.

\paragraph{Mechanism-level span recovery.}
Table~\ref{tab:class-breakdown-span} reports per-class soft span-level recall for GPT-5.4. The proposed taxonomy achieves the highest recall across all 11 mechanism classes. The largest differences over the best alternative are 18.1 percentage points for formal encoding systems, 14.0 points for metaphorical and metonymic encoding, and 9.4 points for referential alias encoding. Results for classes with limited support, particularly cross-linguistic transformation, should be interpreted cautiously.

\paragraph{Compositional ILEs.}

Table~\ref{tab:compositional_app} reports performance on 257 compositional ILE spans across 209 documents. Level~1 evaluates the documents containing compositional ILEs and their evidence spans, whereas Level~2 evaluates the compositional spans directly. At Level~1, the proposed taxonomy exceeds the best alternative by 10.5 percentage points in document accuracy and 4.1 points in span F1. At Level~2, it recovers 147 of 257 spans, compared with 118 under the taxonomy of \citet{leal2024assessing}, corresponding to an 11.3-point recall gain. Its F1 of 0.373 exceeds the highest alternative F1, 0.350 under the taxonomy of \citet{zhang2014appropriate}, by 2.3 points. These results suggest that explicit mechanism guidance aids the recovery of multi-mechanism expressions, although the smaller F1 margin indicates a precision--recall tradeoff.


\clearpage

\begin{table*}[!t]
\centering
\scriptsize
\setlength{\tabcolsep}{1.5pt}
\caption{Cross-lingual soft-span threshold sensitivity at $\tau\in\{0.4,0.5,0.6\}$. The highest value for each model–threshold–metric block is shown in bold.}
\label{tab:threshold-sensitivity-chinese-app}
\begin{tabular*}{\textwidth}{@{\extracolsep{\fill}}llccc|ccc|ccc}
\toprule
\textbf{Model} & \textbf{Variant} & \multicolumn{3}{c|}{$\boldsymbol{\tau=.4}$} & \multicolumn{3}{c|}{$\boldsymbol{\tau=.5}$} & \multicolumn{3}{c}{$\boldsymbol{\tau=.6}$} \\
\cmidrule(lr){3-5} \cmidrule(lr){6-8} \cmidrule(lr){9-11}
 & & Precision & Recall & F1 & Precision & Recall & F1 & Precision & Recall & F1 \\
\midrule
\multirow{6}{*}{GPT-5.4}
 & No Taxonomy & 0.407 & 0.123 & 0.189 & 0.398 & 0.121 & 0.185 & 0.350 & 0.106 & 0.163 \\
 & Zhang et al. & 0.477 & 0.229 & 0.309 & 0.472 & 0.227 & 0.306 & 0.441 & 0.212 & 0.286 \\
 & Calhoun \& Fawcett & \textbf{0.565} & 0.291 & 0.384 & \textbf{0.560} & 0.288 & 0.380 & \textbf{0.526} & 0.271 & 0.358 \\
 & Fillies \& Paschke & 0.450 & 0.298 & 0.359 & 0.450 & 0.298 & 0.359 & 0.428 & 0.283 & 0.341 \\
 & Leal-Arenas \& Corizzo & 0.365 & 0.308 & 0.334 & 0.365 & 0.308 & 0.334 & 0.357 & 0.300 & 0.326 \\
 & Proposed Taxonomy & 0.540 & \textbf{0.362} & \textbf{0.434} & 0.537 & \textbf{0.360} & \textbf{0.431} & 0.500 & \textbf{0.335} & \textbf{0.401} \\
\midrule
\multirow{6}{*}{Sonnet 4.6}
 & No Taxonomy & 0.417 & 0.377 & 0.396 & 0.395 & 0.357 & 0.375 & 0.338 & 0.305 & 0.321 \\
 & Zhang et al. & 0.482 & 0.490 & 0.486 & 0.467 & 0.475 & 0.471 & 0.421 & 0.429 & 0.425 \\
 & Calhoun \& Fawcett & \textbf{0.562} & 0.490 & 0.524 & \textbf{0.545} & 0.475 & 0.508 & \textbf{0.503} & 0.438 & 0.468 \\
 & Fillies \& Paschke & 0.438 & 0.601 & 0.507 & 0.434 & 0.596 & 0.503 & 0.408 & 0.559 & 0.471 \\
 & Leal-Arenas \& Corizzo & 0.387 & 0.547 & 0.454 & 0.382 & 0.539 & 0.447 & 0.366 & 0.517 & 0.429 \\
 & Proposed Taxonomy & 0.462 & \textbf{0.626} & \textbf{0.531} & 0.455 & \textbf{0.616} & \textbf{0.523} & 0.418 & \textbf{0.567} & \textbf{0.481} \\
\midrule
\multirow{6}{*}{DeepSeek V4}
 & No Taxonomy & 0.451 & 0.091 & 0.152 & 0.451 & 0.091 & 0.152 & 0.427 & 0.086 & 0.143 \\
 & Zhang et al. & \textbf{0.618} & 0.219 & 0.324 & \textbf{0.611} & 0.217 & 0.320 & \textbf{0.590} & 0.209 & 0.309 \\
 & Calhoun \& Fawcett & 0.544 & 0.197 & 0.289 & 0.537 & 0.195 & 0.286 & 0.497 & 0.180 & 0.264 \\
 & Fillies \& Paschke & 0.552 & 0.224 & 0.319 & 0.552 & 0.224 & 0.319 & 0.515 & 0.209 & 0.298 \\
 & Leal-Arenas \& Corizzo & 0.531 & 0.232 & 0.322 & 0.520 & 0.227 & 0.316 & 0.497 & 0.217 & 0.302 \\
 & Proposed Taxonomy & 0.594 & \textbf{0.241} & \textbf{0.343} & 0.594 & \textbf{0.241} & \textbf{0.343} & 0.552 & \textbf{0.224} & \textbf{0.319} \\
\bottomrule
\end{tabular*}
\end{table*}

\begin{table*}[!t]
\centering
\scriptsize
\setlength{\tabcolsep}{4pt}
\caption{Per-class soft-span recall for GPT-5.4. Classes follow the proposed taxonomy; $\Delta_{\mathrm{pp}}$ compares the proposed taxonomy and best alternative variants. \textbf{Bold} marks the best value.}
\label{tab:class-breakdown-span}
\begin{tabular}{l c ccccccc}
\toprule
\textbf{Class} & \textbf{Support} & \textbf{Zhang et al.} &
\textbf{Calhoun \& F.} & \textbf{Fillies \& P.} & \textbf{Leal-A. \& C.} &
\textbf{NoTax} & \textbf{Proposed} &
$\boldsymbol{\Delta_{\mathrm{pp}}}$ \\
\midrule
C1--Orthographic Transformation
& 228 & 0.746 & 0.833 & 0.829 & 0.882 & 0.750
& \textbf{0.895} & \textbf{+1.3} \\

C2--Phonetic Substitution
& 365 & 0.595 & 0.603 & 0.630 & 0.655 & 0.592
& \textbf{0.685} & \textbf{+3.0} \\

C3--Formal Compression
& 382 & 0.372 & 0.448 & 0.586 & 0.602 & 0.445
& \textbf{0.615} & \textbf{+1.3} \\

C4--Formal Encoding Systems
& 138 & 0.239 & 0.101 & 0.167 & 0.239 & 0.268
& \textbf{0.449} & \textbf{+18.1} \\

C5--Conventional Sign Reassignment
& 194 & 0.376 & 0.407 & 0.479 & 0.469 & 0.397
& \textbf{0.567} & \textbf{+8.8} \\

C6--Morpho-Lexical Encoding
& 68 & 0.485 & 0.279 & 0.324 & 0.309 & 0.309
& \textbf{0.529} & \textbf{+4.4} \\

C7--Referential Alias Encoding
& 32 & 0.812 & 0.562 & 0.625 & 0.688 & 0.594
& \textbf{0.906} & \textbf{+9.4} \\

C8--Semantic Circumlocution
& 176 & 0.438 & 0.455 & 0.494 & 0.506 & 0.511
& \textbf{0.585} & \textbf{+7.4} \\

C9--Metaphorical and Metonymic Encoding
& 50 & 0.500 & 0.480 & 0.460 & 0.540 & 0.500
& \textbf{0.680} & \textbf{+14.0} \\

C10--Pictorial and Symbolic Encoding
& 91 & 0.297 & 0.429 & 0.560 & 0.473 & 0.484
& \textbf{0.571} & \textbf{+1.1} \\

C11--Cross-Linguistic Transformation
& 13 & 0.231 & 0.000 & 0.077 & 0.077 & 0.077
& \textbf{0.308} & \textbf{+7.7} \\
\bottomrule
\end{tabular}
\end{table*}

\begin{table*}[!t]
\centering
\scriptsize
\setlength{\tabcolsep}{5pt}
\caption{Compositional (multi-class) ILE detection on GPT-5.4. Compositional ILEs are evidence spans annotated with two or more mechanism classes ($n{=}257$ spans across $209$ documents). The best value in each column is \textbf{bold}.}
\label{tab:compositional_app}
\begin{tabular}{lccc|ccc}
\toprule
& \multicolumn{3}{c|}{\textbf{Level 1}} 
& \multicolumn{3}{c}{\textbf{Level 2}} \\
\cmidrule(lr){2-4} \cmidrule(lr){5-7}
\textbf{Variant} 
& \textbf{Document Level} 
& \multicolumn{2}{c|}{\textbf{Span Level}} 
& \multicolumn{3}{c}{\textbf{Span Level}} \\
\cmidrule(lr){2-2} \cmidrule(lr){3-4} \cmidrule(lr){5-7}
& Accuracy & Recall & F1 & Recall & F1 & TP / FN \\
\midrule
No Taxonomy            & 0.555 & 0.475 & 0.608 & 0.393 & 0.320 & 101 / 156 \\
Zhang et al.           & 0.593 & 0.445 & 0.588 & 0.409 & 0.350 & 105 / 152 \\
Calhoun \& Fawcett     & 0.603 & 0.506 & 0.637 & 0.397 & 0.313 & 102 / 155 \\
Fillies \& Paschke     & 0.627 & 0.578 & 0.666 & 0.455 & 0.313 & 117 / 140 \\
Leal-Arenas \& Corizzo & 0.656 & 0.558 & 0.666 & 0.459 & 0.333 & 118 / 139 \\
Proposed Taxonomy      & \textbf{0.761} & \textbf{0.635} & \textbf{0.707} & \textbf{0.572} & \textbf{0.373} & \textbf{147 / 110} \\
\bottomrule
\end{tabular}
\end{table*}

\clearpage

\twocolumn[{
\begin{minipage}{\textwidth}
\centering
\footnotesize
\setlength{\tabcolsep}{6pt}

\captionof{table}{Pairwise differences (proposed $-$ benchmark) on GPT-5.4 using paired bootstrap resampling (10{,}000 iterations). Document-level F1 is macro-averaged, and span-level F1 is soft-span F1.}
\label{tab:pairwise_app}
\vspace{4pt}

\begin{tabular}{lccc|ccc}
\toprule
\textbf{Variant}
& \multicolumn{3}{c|}{\textbf{Accuracy}}
& \multicolumn{3}{c}{\textbf{F1}} \\
& $\Delta_{\mathrm{pp}}$ & 95\% CI & $p$-value
& $\Delta_{\mathrm{pp}}$ & 95\% CI & $p$-value \\
\midrule

\multicolumn{7}{l}{\textit{Document level}} \\

No Taxonomy
& 5.8 & [4.4, 7.3] & $<0.001$
& 6.8 & [5.3, 8.4] & $<0.001$ \\

Zhang et al.
& 7.8 & [6.1, 9.1] & $<0.001$
& 8.7 & [7.2, 10.4] & $<0.001$ \\

Calhoun \& Fawcett
& 7.4 & [5.9, 8.9] & $<0.001$
& 8.6 & [7.1, 10.2] & $<0.001$ \\

Fillies \& Paschke
& 4.9 & [3.5, 6.3] & $<0.001$
& 5.7 & [4.3, 7.2] & $<0.001$ \\

Leal-Arenas \& Corizzo
& 3.8 & [2.4, 5.0] & $<0.001$
& 4.3 & [2.9, 5.7] & $<0.001$ \\

\midrule
\multicolumn{7}{l}{\textit{Span level}} \\

No Taxonomy
& -- & -- & --
& 6.3 & [3.6, 9.1] & $<0.001$ \\

Zhang et al.
& -- & -- & --
& 9.3 & [6.3, 12.9] & $<0.001$ \\

Calhoun \& Fawcett
& -- & -- & --
& 7.3 & [5.2, 9.8] & $<0.001$ \\

Fillies \& Paschke
& -- & -- & --
& 4.1 & [2.2, 6.0] & $<0.001$ \\

Leal-Arenas \& Corizzo
& -- & -- & --
& 2.2 & [0.5, 3.9] & 0.012 \\

\bottomrule
\end{tabular}

\vspace{4pt}
\end{minipage}
}]
\paragraph{Paired-bootstrap comparisons.}
Table~\ref{tab:pairwise_app} reports paired-bootstrap differences with 95\% confidence intervals. At the document level, the proposed taxonomy improves accuracy by 3.8--7.8 percentage points and macro-F1 by 4.3--8.8 points over the alternative variants. All document-level confidence intervals exclude zero, with $p<0.001$. Soft-span F1 gains range from 2.2 to 9.3 points. The smallest span gain is against the best-performing alternative ($\Delta=2.2$ points, 95\% CI $[0.5,3.9]$, $p=0.012$), but remains statistically significant. These comparisons quantify uncertainty for GPT-5.4 and do not imply equivalent effect sizes for the other models.

\paragraph{McNemar comparisons.}
Table~\ref{tab:mcnemar_app} provides a complementary paired comparison of document-level errors. In every comparison, the number of documents correctly classified by the proposed variant but missed by the alternative ($b$) exceeds the reverse count ($c$). The discordant-count ratio ranges from 2.39 for the best-performing alternative ($129/54$) to 4.64 for Zhang et al.\ ($195/42$), with $p<0.001$ in all comparisons. Thus, the accuracy differences consistently favor the proposed taxonomy across paired documents.

\begin{table}[H]
\centering
\scriptsize
\setlength{\tabcolsep}{2pt}
\caption{McNemar's test on document-level accuracy (GPT-5.4). $b/c$: texts correctly classified by the proposed taxonomy but misclassified by the benchmark, versus the reverse.}
\label{tab:mcnemar_app}
\begin{tabular}{lcccc}
\toprule
\textbf{Variant} & \textbf{Accuracy} & \textbf{$\Delta$Acc (pp)} &
\textbf{$b/c$} & \textbf{$p$-value} \\
\midrule
No Taxonomy
& 0.785 & +5.8 & 173 / 57 & $<0.001$ \\
Zhang et al.
& 0.766 & +7.8 & 195 / 42 & $<0.001$ \\
Calhoun \& F.
& 0.769 & +7.4 & 192 / 44 & $<0.001$ \\
Fillies \& P.
& 0.794 & +4.9 & 150 / 52 & $<0.001$ \\
Leal \& C.
& 0.805 & +3.8 & 129 / 54 & $<0.001$ \\
\bottomrule
\end{tabular}
\end{table}

Across the aggregate, class-wise, and paired analyses, the proposed taxonomy improves the recovery of ILE-positive documents and spans, sometimes at a precision cost. Its higher F1 therefore reflects a recall-oriented operating point rather than a universally preferable decision rule. These are detection results: downstream enforcement would require separate assessment of intent, context, severity, fairness, and platform-specific costs.

\clearpage
\onecolumn

\section*{Full Prompt Templates}
The prompt blocks referenced in Appendix~\ref{sec:appendix_b} are reproduced below in full.

\prompttitle{No-Taxonomy Prompt}{app:dev-prompt}

\begin{promptblock}
'''
You are an expert linguist and content annotator specializing in indirect linguistic encoding (ILE) including algospeak, euphemisms, and coded language.
Your task is to identify whether the input text contains ILE. Return a SINGLE JSON object that follows the output schema exactly.

--------
Rules:
--------

DO NOT tag the following as ILE:
- Standard abbreviations commonly used without obfuscation (e.g., "USA", "ASAP")
- Proper names used literally and directly
- Regular typos, misspellings, or spacing errors that are non-intentional
- Literal or descriptive uses of words without indirect meaning


A term may ONLY be annotated as ILE if:

The decoded meaning refers to something that is:
(a) content-sensitive or policy-restricted (including but not limited to sexual, violent, drug-related, self-harm, abusive, harassing, defamatory, or reputational-harm content), or
(b) algorithmically sensitive, moderation-prone, or platform-restricted;

When filling "encoding_evidence":

- Each item MUST be the minimal contiguous euphemistic span, usually a single word, emoji, or very short phrase (e.g., "corn", "unalive", "kermit sewer slide")
- DO NOT return full sentences or long clauses
- The span MUST appear exactly as written in the input text, copied verbatim (same spelling, casing, spacing, punctuation, and emojis)
- Do NOT include surrounding context such as subjects, objects, or full clauses

If one sentence contains multiple ILEs, include each ILE as a separate item.

If NO ILE is present:
- Set encoding_detected to "no"
- Use empty arrays for encoding_evidence and decoded_meaning

------------------------------
Few-Shot Annotation Examples:
------------------------------
Example 1:
Input: {He was caught selling snow behind the club late at night.}
Output: { "analyzed": "yes",  "encoding_detected": "yes",  "encoding_evidence": ["snow"],  "decoded_meaning": ["cocaine"]}

Example 2:
Input: {That site mostly posts corn videos and low-quality memes.}

Output: {  "analyzed": "yes",  "encoding_detected": "yes",  "encoding_evidence": ["corn"],  "decoded_meaning": ["porn"]}

Example 3:
Input: {People keep joking about kermit sewer slide on that platform.}

Output: {  "analyzed": "yes",  "encoding_detected": "yes",  "encoding_evidence": ["kermit sewer slide"],  "decoded_meaning": ["commit suicide"]}

Example 4:
Input: {He felt sad after failing the exam and talked to his friends about it.}

Output: {  "analyzed": "yes",  "encoding_detected": "no",  "encoding_evidence": [],  "decoded_meaning": []}

'''
\end{promptblock}

\prompttitle{Taxonomy-Aware Prompt}{app:tax-prompt}

\begin{promptblock}
'''
SYSTEM_PROMPT = '''
You are an expert linguist and content annotator specializing in indirect linguistic encoding (ILE) including algospeak, euphemisms, and coded language. 

Your task is to identify ILE expressions according to the taxonomy below and return a SINGLE JSON object that follows the output schema exactly.

Use the following categories. The scope of this task is limited to ILE expressions that are instantiated by the categories defined below.

--------------
Categories:
--------------
{Each taxonomy's categories were listed here}

--------------
Rules:
--------------

DO NOT tag the following as ILE:
- Standard abbreviations commonly used without obfuscation (e.g., "USA", "ASAP")
- Proper names used literally and directly
- Regular typos, misspellings, or spacing errors that are non-intentional
- Literal or descriptive uses of words without indirect meaning


A term may ONLY be annotated as ILE if:

The decoded meaning refers to something that is:
(a) content-sensitive or policy-restricted (including but not limited to sexual, violent, drug-related, self-harm, abusive, harassing, defamatory, or reputational-harm content), or
(b) algorithmically sensitive, moderation-prone, or platform-restricted;

Some ILEs may involve multiple transformation or interpretation steps connecting the surface form to the final intended meaning.

For each encoded span:

- include the taxonomy-supported mechanisms that best explain the relationship between the surface form and the final intended meaning,
- preserve mechanism order when sequential interpretation is important,
- and avoid unnecessary or weakly related mechanisms.

If NO ILE is present:
- Set encoding_detected to "no"
- Use empty arrays for encoding_evidence, decoded_meaning, and mechanism

When filling "encoding_evidence":
- Each item MUST be the minimal contiguous ILE span, usually a single word, emoji, or very short phrase (e.g., "corn", "unalive", "kermit sewer slide").
- DO NOT return full sentences or long clauses.
- The span MUST appear exactly as written in the input text, copied verbatim (same spelling, casing, spacing, punctuation, and emojis), with no normalization, correction, or reconstruction.
- If one sentence contains multiple ILEs, include each ILE as a separate item in the array.
- Do NOT include surrounding context such as subjects, objects, or full clauses; only the token(s) that function as the euphemistic or coded expression itself.

When filling "mechanism":
- If NO ILE are detected:
    → "mechanism" MUST be an empty list: []

- If one or more ILEs ARE detected:
    → "mechanism" MUST contain one or more values chosen from the predefined mechanism categories.
    → Each ILE MUST have at least one corresponding mechanism.
    → When multiple ILEs are present, mechanisms SHOULD be aligned by index with encoding_evidence.
    → Do not include extra mechanisms that do not correspond to an ILE.

------------------------------
Few-Shot Annotation Examples:
------------------------------
Example 1:
Input: {He was caught selling snow behind the club late at night.}
Output: { "analyzed": "yes",  "encoding_detected": "yes",  "encoding_evidence": ["snow"],  "decoded_meaning": ["cocaine"],  "mechanism": ["<proper mechanism according to each taxonomy>"]}


Example 2:
Input: {That site mostly posts corn videos and low-quality memes.}

Output: {  "analyzed": "yes",  "encoding_detected": "yes",  "encoding_evidence": ["corn"],  "decoded_meaning": ["porn"],  "mechanism": ["<proper mechanism according to each taxonomy>"]}

Example 3:
Input: {People keep joking about kermit sewer slide on that platform.}

Output: {  "analyzed": "yes",  "encoding_detected": "yes",  "encoding_evidence": ["kermit sewer slide"],  "decoded_meaning": ["commit suicide"],  "mechanism": ["<proper mechanism according to each taxonomy>"]}

Example 4:
Input: {He felt sad after failing the exam and talked to his friends about it.}

Output: {  "analyzed": "yes",  "encoding_detected": "no",  "encoding_evidence": [],  "decoded_meaning": [],  "mechanism": []}

'''
\end{promptblock}

\end{document}